\documentclass[letterpaper]{article} 
\usepackage{aaai2026}  
\usepackage{times}  
\usepackage{helvet}  
\usepackage{courier}  
\usepackage[hyphens]{url}  
\usepackage{graphicx} 
\usepackage{natbib}  
\usepackage{caption} 
\usepackage{algorithm}
\usepackage{algorithmic}
\usepackage{multirow}
\usepackage{pifont}
\usepackage{xcolor}
\usepackage[table,xcdraw]{xcolor}
\usepackage{makecell}
\usepackage{adjustbox}
\usepackage{tabularx}
\usepackage{array}
\usepackage{colortbl,xcolor,array,pifont, arydshln}
\usepackage{subcaption}  
\usepackage{enumitem}
\usepackage[most]{tcolorbox}  
\usepackage{longtable}
\usepackage{ltxtable}
\usepackage{booktabs}

\usepackage{newfloat}
\usepackage{listings}
\DeclareCaptionStyle{ruled}{labelfont=normalfont,labelsep=colon,strut=off} 
\floatstyle{ruled}
\newfloat{listing}{tb}{lst}{}
\floatname{listing}{Listing}
\title{EHRStruct: A Comprehensive Benchmark Framework for Evaluating Large Language Models on Structured Electronic Health Record Tasks}
\author{
    Xiao Yang$^{1}$,
    Xuejiao Zhao$^{2,3}$\thanks{Corresponding Author},
    Zhiqi Shen$^{1}$
}
\affiliations {
    $^1$ College of Computing and Data Science, Nanyang Technological University (NTU), Singapore\\
    $^2$ Joint NTU-UBC Research Centre of Excellence in Active Living for the Elderly (LILY), NTU, Singapore\\
    $^3$ Alibaba-NTU Global e-Sustainability CorpLab (ANGEL), NTU, Singapore\\
    yangxiao.cs@gmail.com, \{xjzhao,zqshen\}@ntu.edu.sg
}

\usepackage{bibentry}

\begin{document}

\maketitle

\begin{abstract}
Structured Electronic Health Record (EHR) data stores patient information in relational tables and plays a central role in clinical decision-making. 
Recent advances have explored the use of large language models (LLMs) to process such data, showing promise across various clinical tasks.
However, the absence of standardized evaluation frameworks and clearly defined tasks makes it difficult to systematically assess and compare LLM performance on structured EHR data.
To address these evaluation challenges, we introduce EHRStruct, a benchmark specifically designed to evaluate LLMs on structured EHR tasks.
EHRStruct defines 11 representative tasks spanning diverse clinical needs and includes 2,200 task-specific evaluation samples derived from two widely used EHR datasets.
We use EHRStruct to evaluate 20 advanced and representative LLMs, covering both general and medical models.
We further analyze key factors influencing model performance, including input formats, few-shot generalisation, and finetuning strategies, and compare results with 11 state-of-the-art LLM-based enhancement
methods for structured data reasoning. 
Our results indicate that many structured EHR tasks place high demands on the understanding and reasoning capabilities of LLMs.
In response, we propose EHRMaster, a code-augmented method that achieves state-of-the-art performance and offers practical insights  to guide future research.
\end{abstract}

\begin{links}
    \link{Code}{https://github.com/YXNTU/EHRStruct}
    \link{Project page}{https://yxntu.github.io/proEHRStruct/}
\end{links}

\section{Introduction}
Structured electronic health records (EHRs)~\cite{hayrinen2008definition} store patient information in relational tables, including  diagnoses, medications, and laboratory results.
Each entry in records corresponds to a specific clinical event or measurement and is often timestamped to capture longitudinal patient trajectories.
In contrast to traditional methods such as SQL-based querying, large language models (LLMs)~\cite{brown2020language} offer greater flexibility, more powerful reasoning abilities, and a natural language interface, and have been increasingly adopted for structured EHR modeling~\cite{li2024scoping}.

The application of LLMs to structured EHR data has recently attracted growing attention, emerging as a promising frontier for clinical AI research.
However, effectively leveraging such data remains challenging even for state-of-the-art LLMs, due to the need for tabular understanding, clinical reasoning, and alignment with user intent~\cite{ren2025comprehensive}.
Recent efforts have attempted to tackle these aspects from different angles:
Yang et al.~\cite{yang2024using} tackle tabular understanding by automating the generation of standardized summary tables from structured clinical trial data.
Zhu et al.~\cite{zhu2024prompting} focus on clinical reasoning tasks, developing a prompting-based approach for predicting outcomes such as mortality, length of stay, and readmission from longitudinal EHR records.
Kwon et al.~\cite{kwon2024ehrcon} address alignment with user intent by verifying the semantic consistency between structured EHR tables and unstructured clinical notes.

Despite promising advances, existing work lacks a unified evaluation framework for assessing LLMs on structured EHR tasks~\cite{lovon2025evaluating}.
This issue manifests in several ways.
First, most studies focus on a limited set of tasks—such as disease prediction~\cite{contreras2024dellirium, hu2024power, hu2024predicting}, mortality risk estimation~\cite{wang2025colacare}, information extraction~\cite{wiest2023text, huang2024critical}, and arithmetic reasoning over tabular data~\cite{yang2024using}—while leaving many clinically important use cases underexplored, including medication recommendation~\cite{thirunavukarasu2023large, shool2025systematic} and clinical named entity recognition~\cite{monajatipoor2024llms, li2024rt}.
Second, even when addressing the same task (e.g., disease prediction), prior studies~\cite{zhu2024prompting, contreras2024dellirium, hu2024power, hu2024predicting} often use different datasets and evaluation protocols, limiting reproducibility and hindering fair model comparison.
Third, there is no consensus on input formatting strategies or experimental setups, leading to inconsistencies in how structured data are presented to LLMs.
Fourth, existing evaluation metrics provide limited interpretability, making it difficult to understand which specific reasoning capabilities contribute to model successes or failures.

\begin{table*}[htbp]
\centering
\begin{tabular}{lllll}
\toprule
\textbf{Task Scenarios} & \textbf{Task Levels} & \textbf{Task Categories} & \textbf{Task IDs} & \textbf{Metrics} \\
\midrule
\multirow{3}{*}[-2pt]{\centering Data-Driven}
  & Understanding
    & {Information retrieval}
      & D-U1/U2
      & Accuracy \\
  \cmidrule(lr){2-5}
  & \multirow{2}{*}[-1pt]{\centering Reasoning}
    & {Data aggregation} & D-R1/R2/R3
      & Accuracy \\
      \cmidrule(lr){3-5}
  & & {Arithmetic computation} & D-R4/R5
      & Accuracy \\
\midrule
\multirow{3}{*}[-1pt]{\centering Knowledge-Driven}
  & Understanding
    & {Clinical identification}
      & K-U1
      & AUC\footnotemark[1] \\
  \cmidrule(lr){2-5}
  & \multirow{2}{*}[-2pt]{\centering Reasoning}
    & {Diagnostic assessment}
      & K-R1/R2
      & AUC \\
      \cmidrule(lr){3-5}
  & & {Treatment planning}
      & K-R3
      & AUC \\
\bottomrule
\end{tabular}
\caption{
Overview of the 11 structured tasks in EHRStruct, categorized by scenario (Data-Driven vs. Knowledge-Driven) and cognitive level (Understanding vs. Reasoning).  
Data-Driven tasks include: 
D-U1/U2: data filtering based on field conditions;  
D-R1/R2/R3: value aggregation including count, average, and sum;  
D-R4/R5: arithmetic reasoning over numeric field trends.  
Knowledge-Driven tasks include:  
K-U1: clinical code identification;  
K-R1: mortality prediction;  
K-R2: disease prediction based on clinical profiles;  
K-R3: medication recommendation tailored to patient characteristics.
}
\label{tab:Tasks_Description}
\end{table*}

To address these limitations, we introduce EHRStruct, a comprehensive benchmark specifically designed to systematically evaluate LLMs on structured EHR tasks. First, to expand task coverage, EHRStruct defines 11 diverse tasks across 6 categories, including information retrieval, data aggregation, and more. These categories are distilled from a thorough analysis of real-world clinical applications and prior research paradigms, ensuring that the selected tasks reflect both operational diversity and clinical relevance. Second, to mitigate dataset inconsistencies, we construct task-specific evaluation samples using standardized data sources from two complementary origins—the synthetic Synthea dataset~\cite{walonoski2018synthea} and the real-world eICU database~\cite{pollard2018eicu}. For each task, we select representative and non-overlapping input–output pairs, validated through cross-checking by multiple domain experts, enabling reproducible and fair comparisons across models. Third, to resolve inconsistencies in input formatting and experimental protocols, we conduct a systematic exploration of input construction strategies and propose a unified framework that clearly distinguishes between different prompt structures for controlled experimentation. Finally, to enhance interpretability and support fine-grained diagnostic analysis, we further classify all tasks along two orthogonal dimensions: evaluation scenario—Data-Driven versus Knowledge-Driven, and cognitive complexity—understanding versus reasoning. This two-dimensional design enables deeper insights into the specific capabilities and limitations of LLMs.

To assess the practicality and effectiveness of our proposed benchmark, EHRStruct, we evaluate 20 representative general and medical LLMs. We primarily focus on zero-shot settings, while also testing 1-shot, 3-shot, and 5-shot configurations to investigate few-shot performance across tasks in our benchmark. In addition, we systematically examine the impact of input formatting, comparing 4 common prompt structures for structured data, and evaluate the effectiveness of task-specific finetuning.
In addition to evaluating base LLMs, we assess 11 representative LLM-based enhancement methods designed to improve structured data performance, including 8 originally developed for non-medical tasks and 3 specifically tailored for medical tasks.
Building on insights from our evaluation of both base LLMs and enhancement methods, we propose EHRMaster—a novel code-augmented framework tailored for structured EHR tasks. EHRMaster operates in three stages: it first generates a high-level solution plan based on the task definition, then aligns key concepts in the plan with relevant data fields, and finally determines whether to generate executable code or proceed with direct reasoning. By structuring execution around task semantics, EHRMaster achieves substantial performance gains across diverse benchmark tasks.
Our main contributions are as follows:
\begin{itemize}
     \item We introduce EHRStruct, a comprehensive benchmark featuring diverse clinically grounded tasks, standardized datasets, systematic input design, and interpretable evaluation to assess LLMs on structured EHR tasks.
    \item We use EHRStruct to systematically evaluate 20 general and medical LLMs and 11 LLM-based enhancement methods, providing extensive analysis and insights into task-specific performance.
    \item We propose EHRMaster, achieving state-of-the-art results on our benchmark and providing valuable inspiration for future research on structured EHR modeling.
\end{itemize}

\begin{figure*}[ht]
  \centering
  \includegraphics[width=\textwidth]{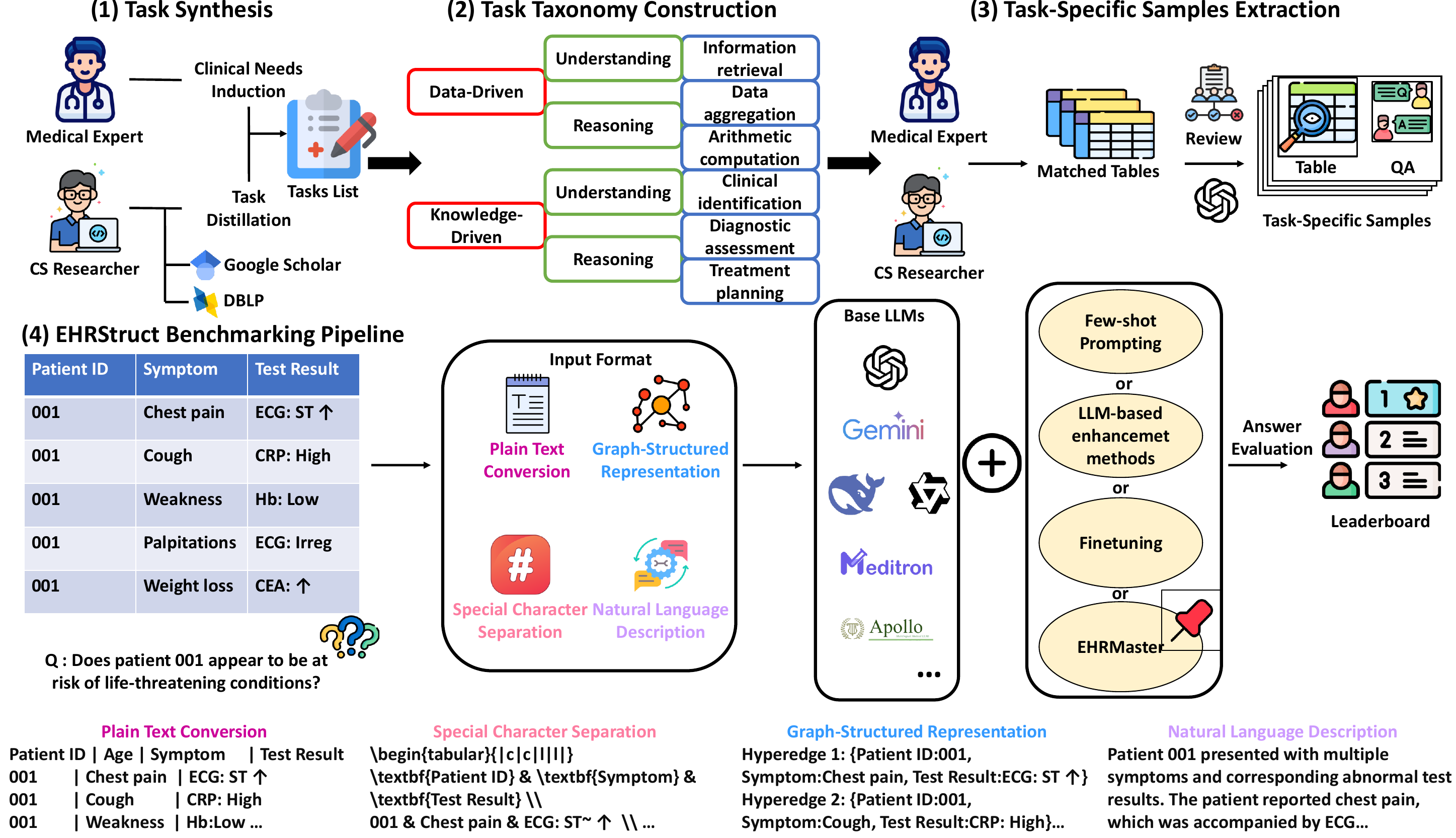}
\caption{Overview of EHRStruct. The figure illustrates the four key components of the benchmark: (1) task synthesis through clinical needs induction and task distillation from prior research; (2) taxonomy construction based on clinical scenarios and reasoning levels; (3) task-specific sample extraction from real and synthetic EHR data; and (4) the model evaluation pipeline, including table input, format conversion, model inference, and answer evaluation.}
  \label{fig:benchmark_overview}
\end{figure*}

\section{Key Findings}
Table~\ref{tab:Tasks_Description} summarizes the benchmark tasks across defined scenarios, levels, and categories, with detailed classification criteria provided in the Appendix Section~\ref{sec:detailed_task_classification}. Our main findings are summarized as follows:
\begin{itemize}
\item \textbf{General LLMs Outperform Medical LLMs:} General LLMs consistently outperform medical models on structured EHR tasks. Among these, closed-source commercial models—especially the Gemini series—achieve the best overall performance.

\item \textbf{LLMs Excel at Data-Driven Tasks:} Overall, LLMs perform better on Data-Driven tasks than on Knowledge-Driven ones.

\item \textbf{Input Format Influences Performance:} Natural language inputs benefit Data-Driven reasoning tasks, while graph-structured prompts help with Data-Driven understanding. No format consistently improves Knowledge-Driven tasks.

\item \textbf{Few-shot Improves Performance:} Few-shot prompting improves performance overall, with 1-shot and 3-shot settings typically outperforming 5-shot.

\item \textbf{Multi-task Fine-tuning Outperforms Single-task Fine-tuning:} While both strategies improve LLM performance, multi-task fine-tuning yields greater gains.

\item \textbf{Enhancement Methods Are Scenario-Specific:} Non-medical enhancement methods underperform in Knowledge-Driven categories, while medical-specific methods struggle in Data-Driven scenarios.
\end{itemize}

\section{EHRStruct} \label{sec:benchmark}
Figure~\ref{fig:benchmark_overview} provides an overview of our benchmark framework, EHRStruct. In this section, we elaborate on its design, covering the data sources, task construction, evaluation setup, and our proposed EHRMaster.

\subsection{Task Synthesis}
Our benchmark defines six categories of tasks that reflect both emerging and established application scenarios in structured EHR modeling. Each task is initially proposed by the CS researcher through task distillation from prior work and existing modeling paradigms, then reviewed and validated by the medical expert to ensure clinical relevance. \textit{Clinical identification} and \textit{Treatment planning} are retained based on expert confirmation, as they are clinically important but remain underexplored in structured settings despite being common in unstructured EHR contexts. Their definitions are customized based on prior work in concept extraction~\cite{ong2023applying, chang2024use} and planning-based dialogue systems~\cite{ullah2024challenges, tan2024medchatzh}. The remaining four task types—\textit{Information retrieval}, \textit{Data aggregation}, \textit{Arithmetic computation}, and \textit{Diagnostic assessment}—are distilled from typical LLM applications to structured EHR data and reflect core reasoning patterns observed in prior studies. Together, these tasks are designed to cover a diverse range of real-world clinical needs in both operational and decision-support scenarios.

\subsection{Task Taxonomy Construction}
As summarized in Table~\ref{tab:Tasks_Description}, we organize the benchmark tasks along three axes: clinical scenario (Data-Driven vs. Knowledge-Driven), cognitive level (Understanding vs. Reasoning), and task category (six functional types). This taxonomy captures both the practical intent and the reasoning complexity of each task, supporting a comprehensive and interpretable evaluation framework. Detailed descriptions are provided in Appendix Section~\ref{sec:detailed_task_description}, and task instructions appear in Appendix Section~\ref{sec:all_instructions}. Such a structured organization also enables fine-grained comparison of model performance across clinically and cognitively diverse settings.

\subsection{Task-Specific Samples Extraction}

Our benchmark is built from two representative structured EHR sources. The first is \textbf{Synthea}~\cite{walonoski2018synthea}\footnote{\url{https://github.com/synthetichealth/synthea}}, a synthetic dataset simulating realistic patient records without privacy concerns. The second is the \textbf{eICU Collaborative Research Database}~\cite{pollard2018eicu}\footnote{\url{https://eicu-crd.mit.edu/}}, a real-world ICU dataset comprising clinically rich, multi-institutional structured tables. Together, they ensure coverage of both simulated and authentic clinical scenarios.

For each task,  the  CS researcher and the medical expert jointly screen and identify the most relevant tables based on the task definition and schema content. Once matched, we construct 100 evaluation samples per task per dataset, resulting in 2,200 annotated instances across 11 tasks. Representative data rows are selected to ensure clinical diversity, and GPT-4o is used to generate question--answer pairs conditioned on the task definition, table schema, and sampled content. All outputs undergo two-stage validation: medical reviewers assess the correctness and plausibility of answers, while technical reviewers verify that each question is faithful to the task objectives and input semantics.

\begin{table}[ht]
\centering
\begin{tabular}{@{}llc@{}}
\toprule
\textbf{Types} & \textbf{Models} & \textbf{\# Params} \\ \midrule

\multirow{11}{*}{\rotatebox{90}{\shortstack{General\\Large Language Models}}} & GPT-3.5 Turbo~\cite{openai2023} & Commercial \\
 & GPT-4.1~\cite{openai2025} & Commercial \\ \cmidrule(lr){2-3}
 & Gemini 1.5~\cite{deepmind2024a} & Commercial \\
 & Gemini 2.0~\cite{deepmind2024b} & Commercial \\
 & Gemini 2.5~\cite{deepmind2025} & Commercial \\ \cmidrule(lr){2-3}
 & DeepSeek-V2.5~\cite{liu2024deepseeka} & 236B \\
 & DeepSeek-V3~\cite{liu2024deepseekb} & 685B \\ \cmidrule(lr){2-3}
 & Qwen-7B~\cite{team2024qwen2} & 7B \\
& Qwen-14B~\cite{team2024qwen2} & 14B \\
 & Qwen-32B~\cite{team2024qwen2} & 32B \\
 & Qwen-72B~\cite{team2024qwen2} & 72B \\ \midrule

\multirow{9}{*}{\rotatebox{90}{\shortstack{Medical\\Large Language Models}}} & Huatuo~\cite{zhang2023huatuogpt} & 7B \\
 & HEAL~\cite{han2023medalpaca} & 7B \\
 & Meditron-7B~\cite{chen2023meditron} & 7B \\ \cmidrule(lr){2-3}
 & MedAlpaca-13B~\cite{yuan2024continued} & 13B \\
 & JMLR~\cite{wang2024jmlr} & 13B \\
 & PMC\_LLaMA\_13B~\cite{wu2024pmc} & 13B \\ \cmidrule(lr){2-3}
 & Med42-70B~\cite{christophe2024med42} & 70B \\
 & Apollo~\cite{wang2024apollo} & 70B \\
 & CancerLLM~\cite{li2024cancerllm} & 70B \\ \bottomrule
\end{tabular}
\caption{List of 20 LLMs, including 11 general and 9 medical models, covering both open-source and commercial releases with parameter sizes ranging from 7B to 685B.}
\label{table:modeldescription}
\end{table}

\subsection{Evaluation Setup}
We evaluate a diverse set of large language models (LLMs) on our benchmark, including both general-purpose and medical-domain models. Table~\ref{table:modeldescription} summarizes the 20 models included in our study, detailing their model types and parameter sizes. For each task ID, we evaluate all 20 LLMs using 200 question–answer pairs (100 from the synthetic Synthea dataset and 100 from the real-world eICU dataset), ensuring balanced coverage across data sources. We employ 4 distinct formats to transform structured EHR data into text inputs and report results separately for each data source. All evaluations use single-turn generation with consistent decoding parameters (e.g., temperature and maximum token limits) to ensure fair model comparisons.

Beyond these benchmark-wide evaluations, we conduct detailed few-shot and fine-tuning experiments on the Gemini series to investigate their potential on structured EHR tasks. We also reproduce 11 existing methods for structured data reasoning—8 from other domains and 3 from clinical settings—and systematically evaluate their performance on our benchmark for comparison. Finally, we evaluate our proposed EHRMaster method on the benchmark to demonstrate its effectiveness relative to both general-purpose LLMs and existing structured-data reasoning approaches.

\subsection{Task Metrics}
For K-U and K-R tasks, we adopt the Area Under the ROC Curve (AUC) as the main evaluation metric.  
Although the model outputs are discrete class labels rather than continuous probabilities, AUC remains valid in this setting.  
When predictions are binary, AUC is mathematically equivalent to balanced accuracy, which measures the mean of sensitivity and specificity.  
We prefer AUC for K-U and K-R tasks because medical datasets often exhibit substantial class imbalance, and AUC provides a more robust evaluation of model discrimination under such conditions.

\subsection{EHRMaster Overview}
Here, we briefly outline our proposed EHRMaster framework. A detailed description is provided in Appendix Section~\ref{sec:EHRMaster}. EHRMaster operates in three key stages:

\textbf{Solution Planning: }Generates a high-level solution plan based on the question, outlining the reasoning steps required to find the answer. 

For example, given the task  ``calculate the total cost of a patient's hospital stay'' , EHRMaster generates a plan with the following steps: (1) locate the admission and discharge events; (2) retrieve billing entries linked to the hospitalization; and (3) compute the total cost. The plan is expressed in natural language and serves as a blueprint for downstream concept alignment and execution.

\textbf{Concept Alignment: }Maps the abstract concepts from the solution plan to the corresponding fields in the structured data, ensuring alignment with the task requirements. 

For example, in the hospital cost estimation task, ``admission time'' and ``discharge time'' are linked to the \texttt{ADMITTIME} and \texttt{DISCHTIME} fields in the \texttt{ADMISSIONS} table, while ``billing entries'' are mapped to the \texttt{ITEM\_COST} field in the \texttt{BILLING} table.

\begin{table*}[htbp]
\centering

\resizebox{0.89\textwidth}{!}{%
\begin{tabular}{ll*{7}{c}*{5}{c}}
  \toprule
  \multirow{4}{*}{\bfseries Types} & \multirow{4}{*}{\bfseries Models} 
  & \multicolumn{7}{c}{\bfseries Data-Driven } 
  & \multicolumn{5}{c}{\bfseries Knowledge-Driven} \\
  \cmidrule(lr){3-9} \cmidrule(lr){10-13}
  & & \multicolumn{2}{c}{\bfseries U (\,\%)} & \multicolumn{5}{c}{\bfseries R (\,\%)} 
  & \multicolumn{1}{c}{\bfseries U (\,\%)} & \multicolumn{3}{c}{\bfseries R (\,\%)} \\
  & & D-U1 & D-U2 & D-R1 & D-R2 & D-R3 & D-R4 & D-R5 
  & K-U1  & K-R1 & K-R2 & K-R3 \\
  \cmidrule(lr){3-14}
  & & ACC & ACC & ACC & ACC & ACC & ACC & ACC 
  & AUC & AUC & AUC & AUC  \\
  \midrule
  \multirow{10}{*}{\rotatebox{90}{\bfseries General LLMs}} & GPT-3.5 Turbo & 6 & 15 & 14 & 18 & 7 & 7 & 24 
  & \textcolor{red}{\ding{55}}  & {\color[HTML]{32CB00}\textbf{58.1}} & {\color[HTML]{F56B00}\textbf{55.4}} & {\color[HTML]{3166FF}\textbf{52.9}} \\
  & GPT-4.1 & {\color[HTML]{F56B00} \textbf{79}} & {\color[HTML]{F56B00} \textbf{51}} & {\color[HTML]{F56B00} \textbf{52}} & {\color[HTML]{F56B00} \textbf{56}} & {\color[HTML]{F56B00} \textbf{48}} & {\color[HTML]{32CB00} \textbf{70}} & {\color[HTML]{32CB00} \textbf{84}}
  & {\color[HTML]{3166FF}\textbf{55}} & 55.6 & 53.2 & 51 \\
    \cmidrule(lr){2-13}
  & Gemini 1.5 & 29 & 34 & {\color[HTML]{32CB00} \textbf{32}} & 41 & 21 & 19 & 16 
  & \textcolor{red}{\ding{55}} & 55.6 & \textcolor{red}{\ding{55}} & \textcolor{red}{\ding{55}} \\
  & Gemini-2.0 & 64 & {\color[HTML]{32CB00} \textbf{43}} & 21 & 30 & {\color[HTML]{32CB00} \textbf{24}} & 54 & 67 
  & {\color[HTML]{F56B00}\textbf{52}}  & 57.7 & {\color[HTML]{3166FF}\textbf{56.2}} & {\color[HTML]{32CB00}\textbf{51.6}} \\
  & Gemini 2.5 & {\color[HTML]{3166FF} \textbf{98}} & {\color[HTML]{3166FF} \textbf{58}} & {\color[HTML]{3166FF} \textbf{92}} & {\color[HTML]{3166FF} \textbf{82}} & {\color[HTML]{3166FF} \textbf{83}} & \textcolor[HTML]{3166FF}{\ding{51}} & \textcolor[HTML]{3166FF}{\ding{51}} 
  & \textcolor{red}{\ding{55}} & {\color[HTML]{3166FF}\textbf{58.7}} & {\color[HTML]{32CB00}\textbf{54.1}} & \textcolor{red}{\ding{55}} \\
    \cmidrule(lr){2-13}
  & DeepSeek-V2.5 & {\color[HTML]{32CB00} \textbf{72}} & 41 & 18 & 51 & 14 & 44 & 52 
  & {\color[HTML]{32CB00}\textbf{51}} & \textcolor{red}{\ding{55}} & \textcolor{red}{\ding{55}} & \textcolor{red}{\ding{55}} \\
  & DeepSeek-V3 & {\color[HTML]{32CB00} \textbf{72}} & 41 & 8 & 37 & 12 & {\color[HTML]{F56B00} \textbf{72}} & {\color[HTML]{F56B00} \textbf{90}} 
  & \textcolor{red}{\ding{55}}  & 52.8 & \textcolor{red}{\ding{55}} & \textcolor{red}{\ding{55}} \\
    \cmidrule(lr){2-13}
  & Qwen-7B & 1 & 7 & 4 & 24 & 1 & \textcolor{red}{\ding{55}} & \textcolor{red}{\ding{55}} 
  & \textcolor{red}{\ding{55}} & \textcolor{red}{\ding{55}} & \textcolor{red}{\ding{55}} & \textcolor{red}{\ding{55}} \\
  & Qwen-14B & 4 & 30 & 19 & 17 & 11 & 16 & 4 
  & \textcolor{red}{\ding{55}}  & \textcolor{red}{\ding{55}} & \textcolor{red}{\ding{55}} & \textcolor{red}{\ding{55}} \\
  & Qwen-32B & 25 & 25 & 24 & 26 & 15 & 47 & 10 
  & \textcolor{red}{\ding{55}}& {\color[HTML]{F56B00}\textbf{58.3}} & 51 & \textcolor{red}{\ding{55}} \\
  & Qwen-72B & 15 & 6 & 27 & {\color[HTML]{32CB00} \textbf{48}} & 20 & 41 & 29 
  & \textcolor{red}{\ding{55}}  & \textcolor{red}{\ding{55}} & \textcolor{red}{\ding{55}} & {\color[HTML]{F56B00}\textbf{52.2}} \\
  \midrule
  \multirow{9}{*}{\rotatebox{90}{\bfseries Medical LLMs}} & Huatuo & \textcolor{red}{\ding{55}} & \textcolor{red}{\ding{55}} & \textcolor{red}{\ding{55}} & \textcolor{red}{\ding{55}} & \textcolor{red}{\ding{55}} & \textcolor{red}{\ding{55}} & \textcolor{red}{\ding{55}} 
  & \textcolor{red}{\ding{55}}  & \textcolor{red}{\ding{55}} & \textcolor{red}{\ding{55}} & \textcolor{red}{\ding{55}} \\
  & HEAL & \textcolor{red}{\ding{55}} & \textcolor{red}{\ding{55}} & 1 & 8 & \textcolor{red}{\ding{55}} & \textcolor{red}{\ding{55}} & \textcolor{red}{\ding{55}} 
  & \textcolor{red}{\ding{55}}  & \textcolor{red}{\ding{55}} & \textcolor{red}{\ding{55}} & \textcolor{red}{\ding{55}} \\
  & Meditron-7B & \textcolor{red}{\ding{55}} & 3 & \textcolor{red}{\ding{55}} & 6 & \textcolor{red}{\ding{55}} & \textcolor{red}{\ding{55}} & \textcolor{red}{\ding{55}} 
  & \textcolor{red}{\ding{55}} & \textcolor{red}{\ding{55}} & \textcolor{red}{\ding{55}} & \textcolor{red}{\ding{55}} \\
    \cmidrule(lr){2-13}
  & MedAlpaca-13B & 2 & 11 & 6 & 4 & 2 & 10 & \textcolor{red}{\ding{55}} 
  & \textcolor{red}{\ding{55}}  & \textcolor{red}{\ding{55}} & \textcolor{red}{\ding{55}} & \textcolor{red}{\ding{55}} \\
  & JMLR & 1 & 3 & 11 & 10 & 6 & 7 & 3 
  & \textcolor{red}{\ding{55}}  & \textcolor{red}{\ding{55}} & \textcolor{red}{\ding{55}} & \textcolor{red}{\ding{55}} \\
  & PMC\_LLaMA\_13B & 6 & 6 & 15 & 13 & 10 & 8 & \textcolor{red}{\ding{55}} 
  & \textcolor{red}{\ding{55}}  & \textcolor{red}{\ding{55}} & \textcolor{red}{\ding{55}} & \textcolor{red}{\ding{55}} \\
    \cmidrule(lr){2-13}
  & Med42-70B & 13 & 3 & 18 & 17 & 11 & 27 & 18 
  & \textcolor{red}{\ding{55}}  & \textcolor{red}{\ding{55}} & \textcolor{red}{\ding{55}} & \textcolor{red}{\ding{55}} \\
  & Apollo & 11 & 5 & 17 & 12 & 6 & 20 & 11 
  & \textcolor{red}{\ding{55}} & \textcolor{red}{\ding{55}} & \textcolor{red}{\ding{55}} & \textcolor{red}{\ding{55}} \\
  & CancerLLM & 10 & 16 & 20 & 28 & 15 & 33 & 25 
  & \textcolor{red}{\ding{55}}  & \textcolor{red}{\ding{55}} & \textcolor{red}{\ding{55}} & \textcolor{red}{\ding{55}} \\
  \bottomrule
\end{tabular}
}
\caption{
Performance of LLMs on Structured EHR Tasks under the zero-shot setting(Synthea).
\textcolor{red}{\ding{55}} indicates no valid output.
\textcolor[HTML]{3166FF}{\ding{51}} indicates a perfect score of 100.
\textbf{\textcolor[HTML]{3166FF}{1\textsuperscript{st}}}, \textbf{\textcolor[HTML]{F56B00}{2\textsuperscript{nd}}}, and \textbf{\textcolor[HTML]{32CB00}{3\textsuperscript{rd}}} denote the best, second‑best, and third‑best results, respectively.
}
\label{tab:synthea_benchmark}
\end{table*}

\textbf{Adaptive Execution: } Determines whether the question can be addressed using code-based execution. If applicable, it generates executable code and uses a code interpreter to obtain the result; otherwise, it proceeds with direct reasoning over the data. 

For example, in the hospital cost estimation task, once the relevant fields have been identified, EHRMaster  generate Python code to filter billing entries by admission and discharge timestamps and compute the total cost. In contrast, for tasks like assessing treatment effectiveness based on heterogeneous clinical events, EHRMaster may bypass code and instead use multi-step language reasoning grounded in the aligned data.
\section{Results}

In this section, we provide an overview of the results, analyses, and experimental findings for the LLMs evaluated in our benchmark. Detailed results and further analyses are presented in Appendix Section~\ref{sec:detailed_results}.

\subsection{Overall Benchmark Results}\label{sec:mainresults}
Table~\ref{tab:synthea_benchmark} reports the performance of all evaluated models on the synthetic Synthea dataset. We organize results by task scenario—Data-Driven and Knowledge-Driven—and further distinguish tasks by level, specifically Understanding (U) and Reasoning (R). Although the specific question–answer pairs differ across datasets, we observe highly consistent performance patterns on the real-world eICU dataset as well, which is reported in Appendix Table~\ref{tab:eicu_benchmark}.

First, general LLMs consistently outperform medical LLMs across nearly all task categories. This performance gap is particularly evident in the Knowledge-Driven scenario, where medical models frequently fail to produce valid outputs or achieve meaningful accuracy or AUC scores. Notably, none of the medical models rank among the top three performers for any task ID. In contrast, closed-source commercial models—especially the Gemini series—achieve the highest overall performance, demonstrating robust generalization across synthetic structured EHR tasks. These results suggest that general models benefit from broad pretraining on diverse text sources, which may indirectly support structured data understanding even without explicit domain-specific adaptation.

\begin{figure*}[ht]
  \centering
  \begin{subfigure}[b]{0.33\textwidth}
    \centering
    \includegraphics[width=\linewidth]{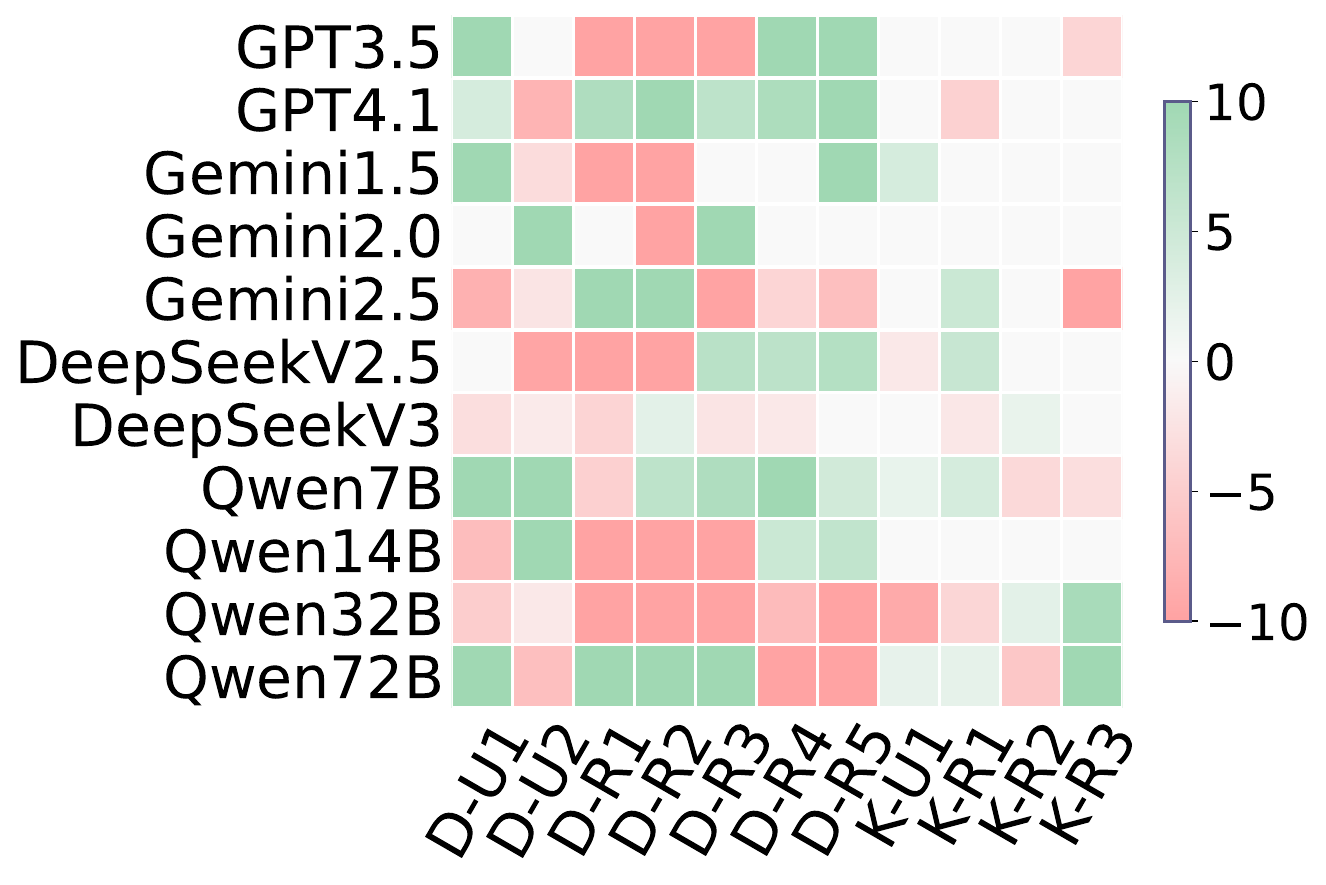}
    \caption*{special character separation}
  \end{subfigure}
  \hfill
  \begin{subfigure}[b]{0.33\textwidth}
    \centering
    \includegraphics[width=\linewidth]{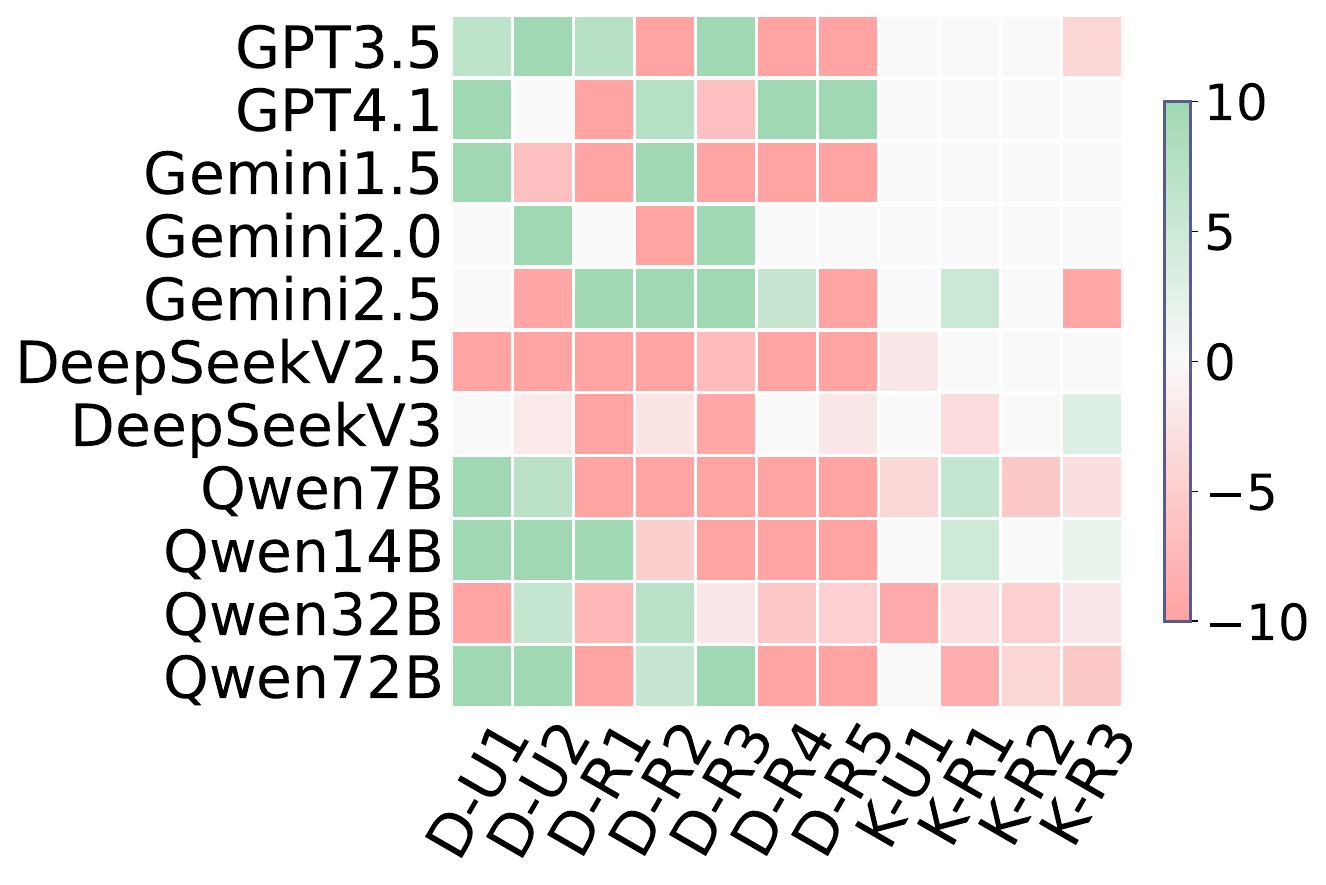}
    \caption*{graph structured representation}
  \end{subfigure}
  \hfill
  \begin{subfigure}[b]{0.33\textwidth}
    \centering
    \includegraphics[width=\linewidth]{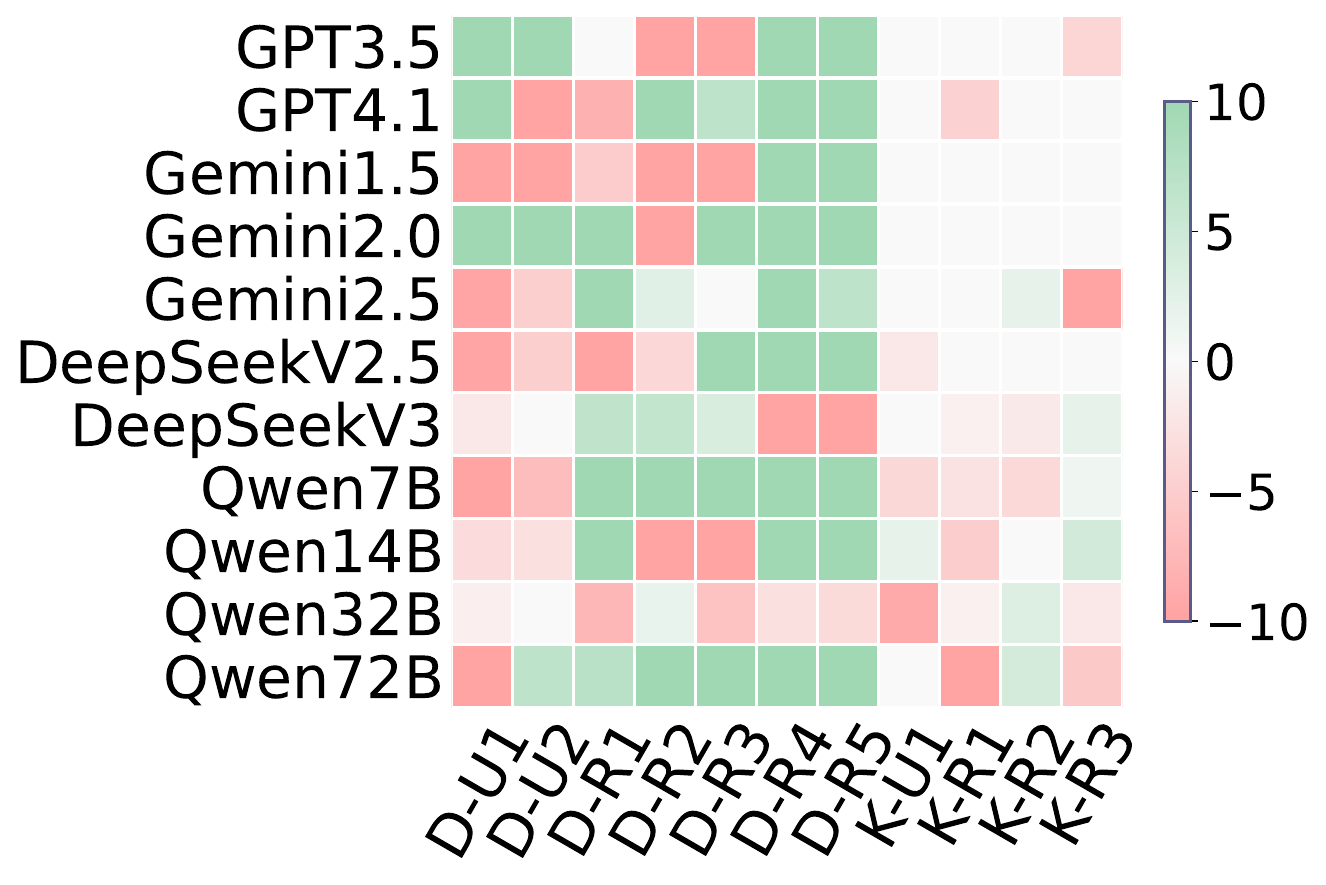}
    \caption*{natural language description}
  \end{subfigure}
  \caption{Relative Performance Gains from Different input formats across LLMs.}
  \label{fig:formatting_comparison}
\end{figure*}

\begin{figure}[htbp]
    \centering
    \begin{subfigure}[t]{0.3\columnwidth}
        \includegraphics[width=\linewidth]{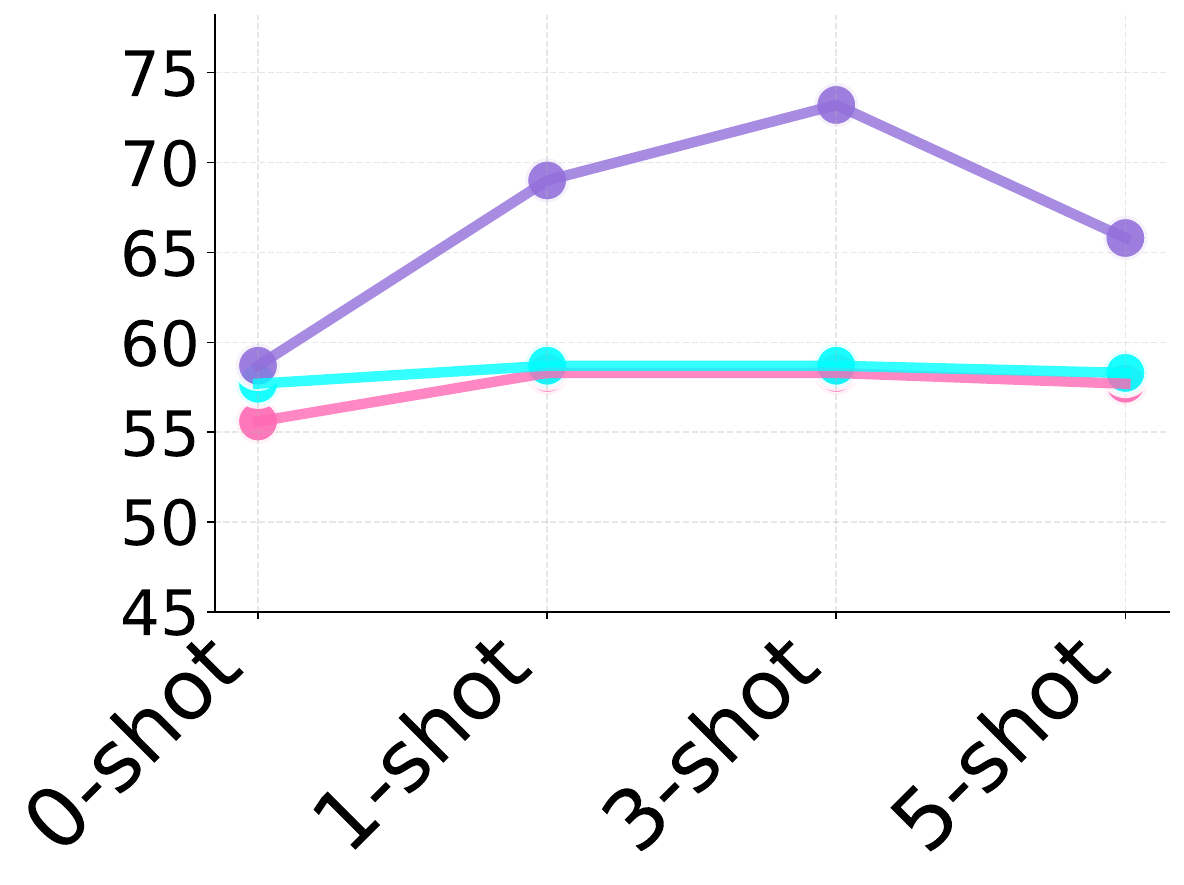}
        \caption{K-R1}
    \end{subfigure}
    \hfill
    \begin{subfigure}[t]{0.3\columnwidth}
        \includegraphics[width=\linewidth]{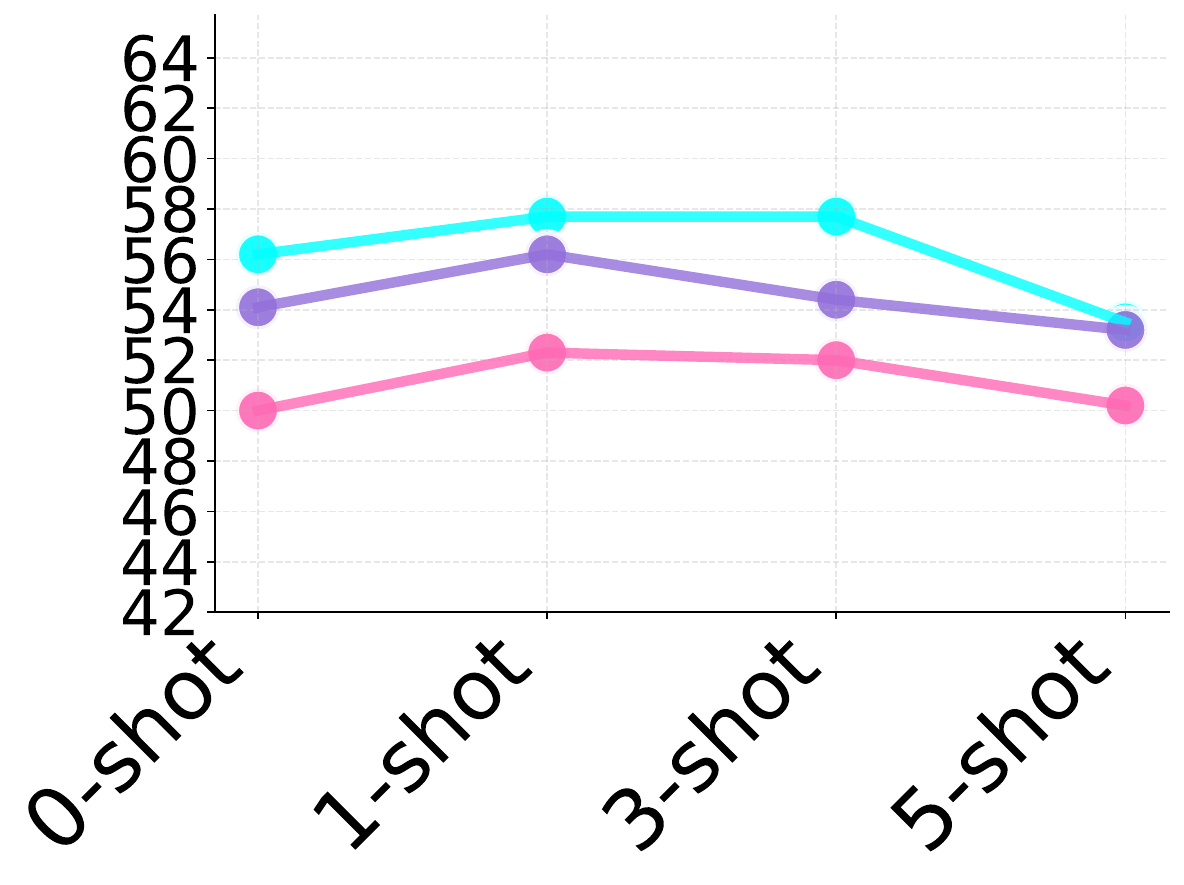}
        \caption{K-R2}
    \end{subfigure}
    \hfill
    \begin{subfigure}[t]{0.3\columnwidth}
        \includegraphics[width=\linewidth]{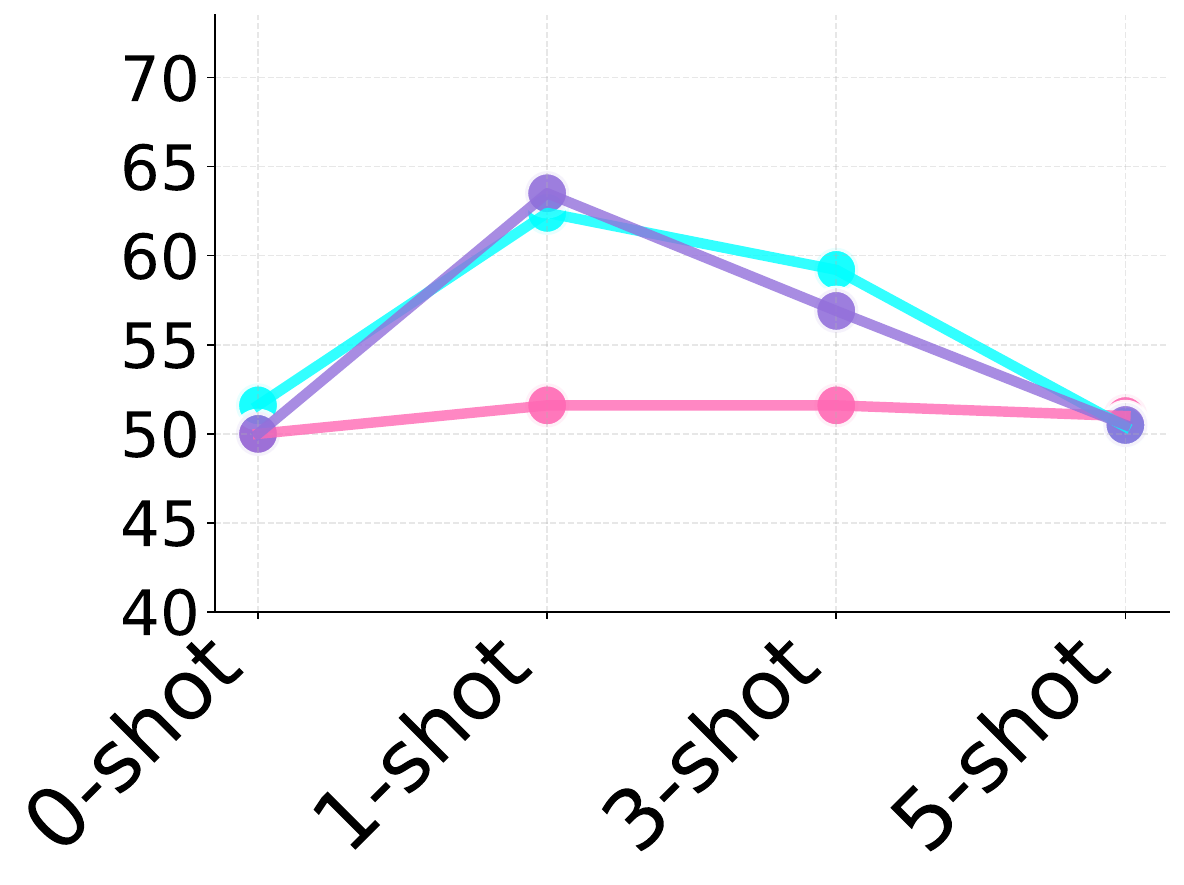}
        \caption{K-R3}
    \end{subfigure}
\raisebox{0.5ex}{\textcolor[HTML]{FF69B4}{\rule{0.8cm}{1pt}}} Gemini 1.5 \quad
\raisebox{0.5ex}{\textcolor[HTML]{00FFFF}{\rule{0.8cm}{1pt}}} Gemini-2.0 \quad
\raisebox{0.5ex}{\textcolor[HTML]{9370DB}{\rule{0.8cm}{1pt}}} Gemini-2.5

    \caption{Performance of representative LLMs on two Scenarios under few-shot (1, 3, and 5-shot) learning settings.}
    \label{fig:fewshot_selected_tasks}
\end{figure}

Second, model performance shows clear variation across task scenarios and levels, reflecting the different demands posed by each task category. In the Data-Driven scenario, strong general-purpose LLMs perform well on both understanding and reasoning tasks, suggesting that, when structured inputs are properly formatted, general models can interpret and reason over structured data with reasonable accuracy. In contrast, Knowledge-Driven tasks present substantially greater challenges. For the understanding task of Clinical Code Mapping (K-U1), general models achieve only moderate AUC scores (typically between 50–60\%), while most medical-domain models fail to produce valid outputs. Performance drops even further on reasoning tasks such as Diagnostic Assessment (K-R1, K-R2) and Treatment Planning (K-R3), where many models struggle to generate meaningful predictions. This gap highlights the difficulty of integrating external clinical knowledge into structured data interpretation and underscores the need for models that better handle medical semantics and complex reasoning.

\subsection{Few-shot Analysis} \label{sec:fewshot}
Figure~\ref{fig:fewshot_selected_tasks} presents few-shot performance curves for a subset of Knowledge-Driven tasks, with full results provided in Appendix Figure~\ref{fig:fewshot_detailed}. Overall, few-shot prompting is beneficial, with 1-shot and 3-shot settings typically outperforming 5-shot. Compared to Gemini 1.5, Gemini 2.0 and 2.5 exhibit more pronounced performance gains under few-shot settings, suggesting a greater sensitivity to in-context examples and a stronger ability to generalize from limited demonstrations.

\subsection{Effect of Input Formats} \label{sec:inputprompt}
We examine 4 input formats for converting structured electronic health record data into text: plain text conversion, special character separation, graph-structured representation, and natural language description. These formats are used to test how different input styles affect model performance on our benchmark.

\begin{figure}[htbp]
    \centering
    \begin{subfigure}[t]{0.48\columnwidth}
        \centering
        \includegraphics[width=\linewidth]{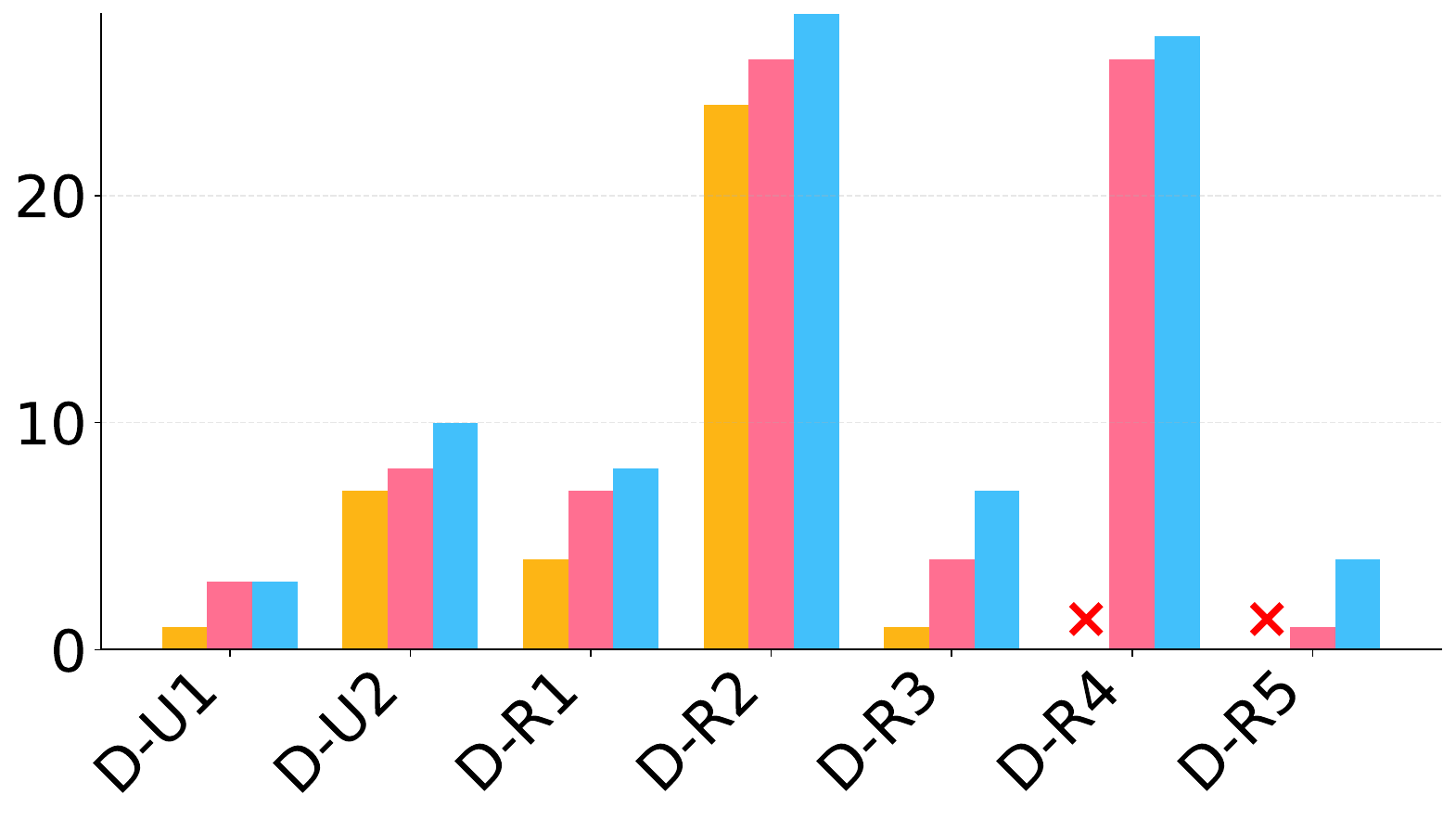}
        \caption{Data-Driven}
        \label{fig:d_finetune}
    \end{subfigure}
    \hfill
    \begin{subfigure}[t]{0.48\columnwidth}
        \centering
        \includegraphics[width=\linewidth]{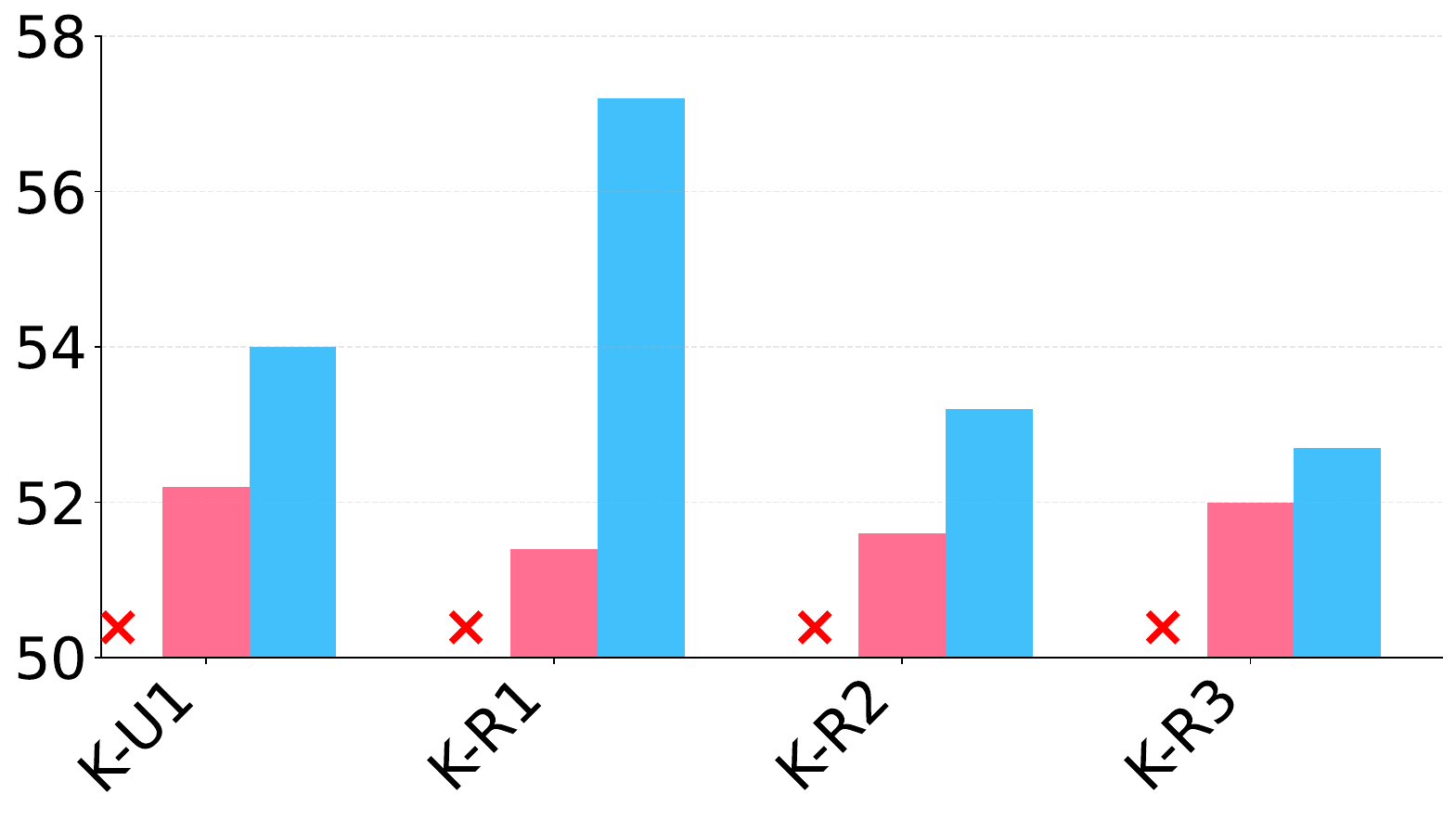}
        \caption{Knowledge-Driven}
        \label{fig:k_finetune}
    \end{subfigure}
\begin{minipage}{\linewidth}
    \centering
    \begin{tabular}{ccc}
        {\color[HTML]{FDB515}\rule{10pt}{6pt}}~Baseline &
        {\color[HTML]{FF6F91}\rule{10pt}{6pt}}~Single &
        {\color[HTML]{42C0FB}\rule{10pt}{6pt}}~Multi
    \end{tabular}
\end{minipage}
    \caption{
        Finetuning results on all \textbf{targeted categories}. Single-task indicates separate finetuning on each task; multi-task indicates joint finetuning across all tasks.
    }
    \label{fig:finetune_all}
\end{figure}
As shown in Figure~\ref{fig:formatting_comparison}, input format has a clear impact on performance. Natural language description improves results on Data-Driven reasoning tasks, especially for strong models such as the Gemini and GPT series. Graph-structured input is more effective for Data-Driven understanding tasks. Across all Knowledge-Driven tasks, however, no format yields consistent improvement. This shows that input format helps in certain settings, but deeper modeling efforts are needed for tasks requiring clinical knowledge.

\subsection{Finetuning Analysis} \label{sec:finetune}
To analyze the fine-tuning performance of large language models on structured EHR tasks, we construct additional training samples specifically for fine-tuning, which are completely separate from the evaluation samples used in benchmarking. Each fine-tuning dataset contains 30 task-specific question–answer table pairs following the same instruction format as in the evaluation, ensuring consistency of task structure while preventing data leakage. We adopt two fine-tuning strategies: single-task fine-tuning trains each task separately, while multi-task fine-tuning combines all tasks into one set. All experiments are conducted on the Qwen-7B model using LoRA-based parameter-efficient fine-tuning with a 10\% validation split, a learning rate of 0.0001, three training epochs, a batch size of 8, and LoRA hyperparameters set to rank 8, alpha 32, and dropout 0.05.

Figure~\ref{fig:finetune_all} shows that fine-tuning substantially improves model performance across both Data-Driven and Knowledge-Driven tasks. Across all tasks, multi-task fine-tuning consistently outperforms single-task fine-tuning, likely because joint training helps the model learn shared structures and reasoning patterns relevant to structured EHR data.

\begin{figure*}[htbp]
      \centering
      \begin{subfigure}[b]{\textwidth}
        \centering
        \begin{subfigure}[b]{0.32\textwidth}
          \includegraphics[width=0.49\textwidth]{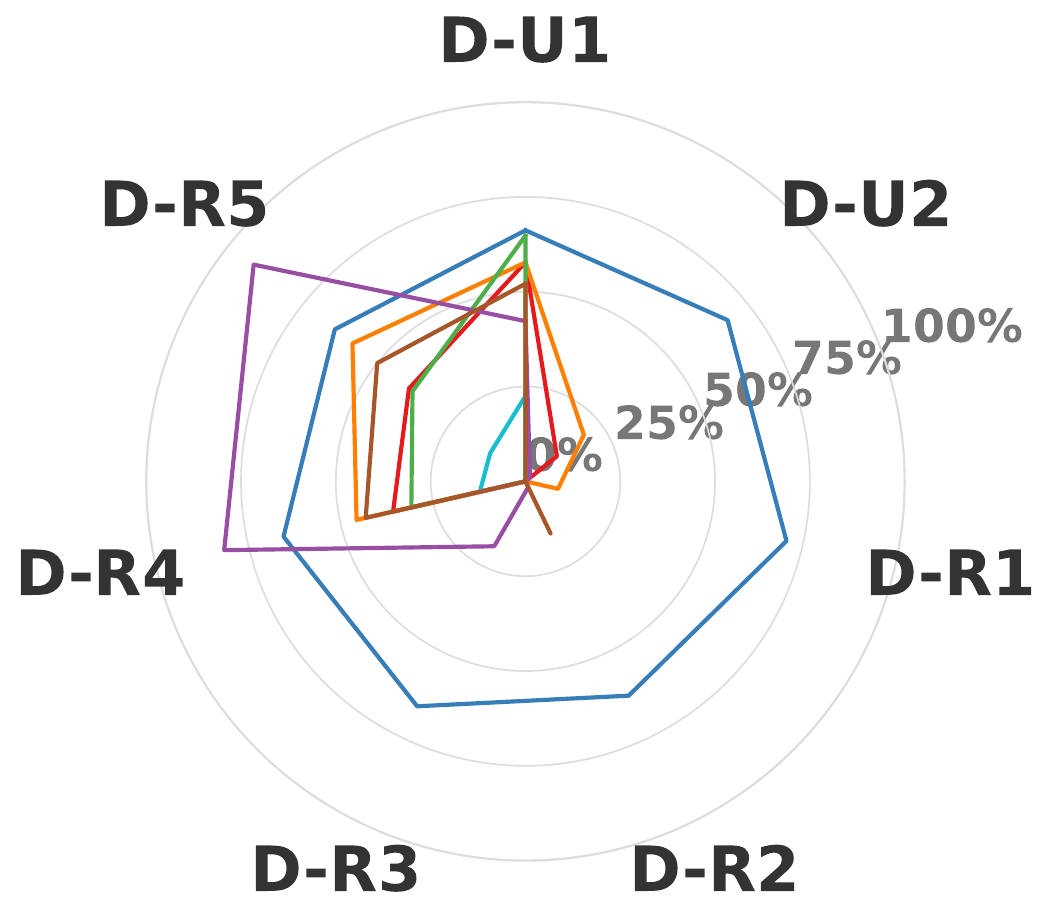}
          \includegraphics[width=0.49\textwidth]{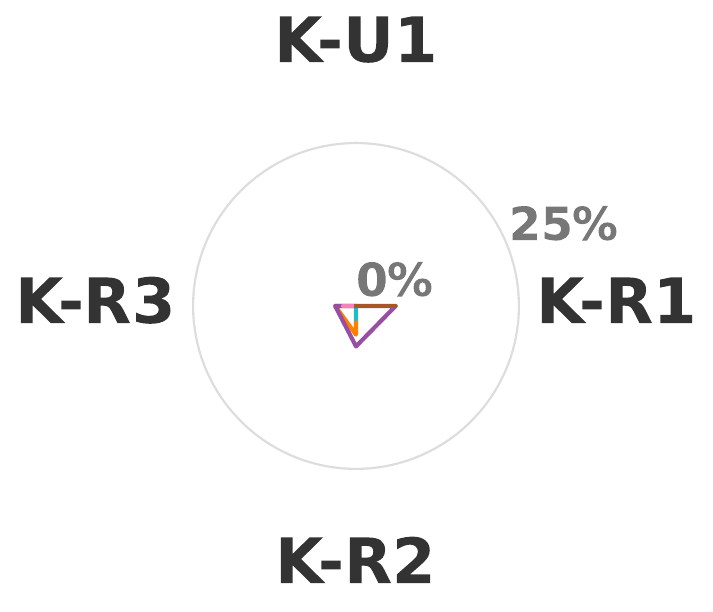}
          \caption{Gemini-1.5}
        \end{subfigure}
        \hfill
        \begin{subfigure}[b]{0.32\textwidth}
          \includegraphics[width=0.49\textwidth]{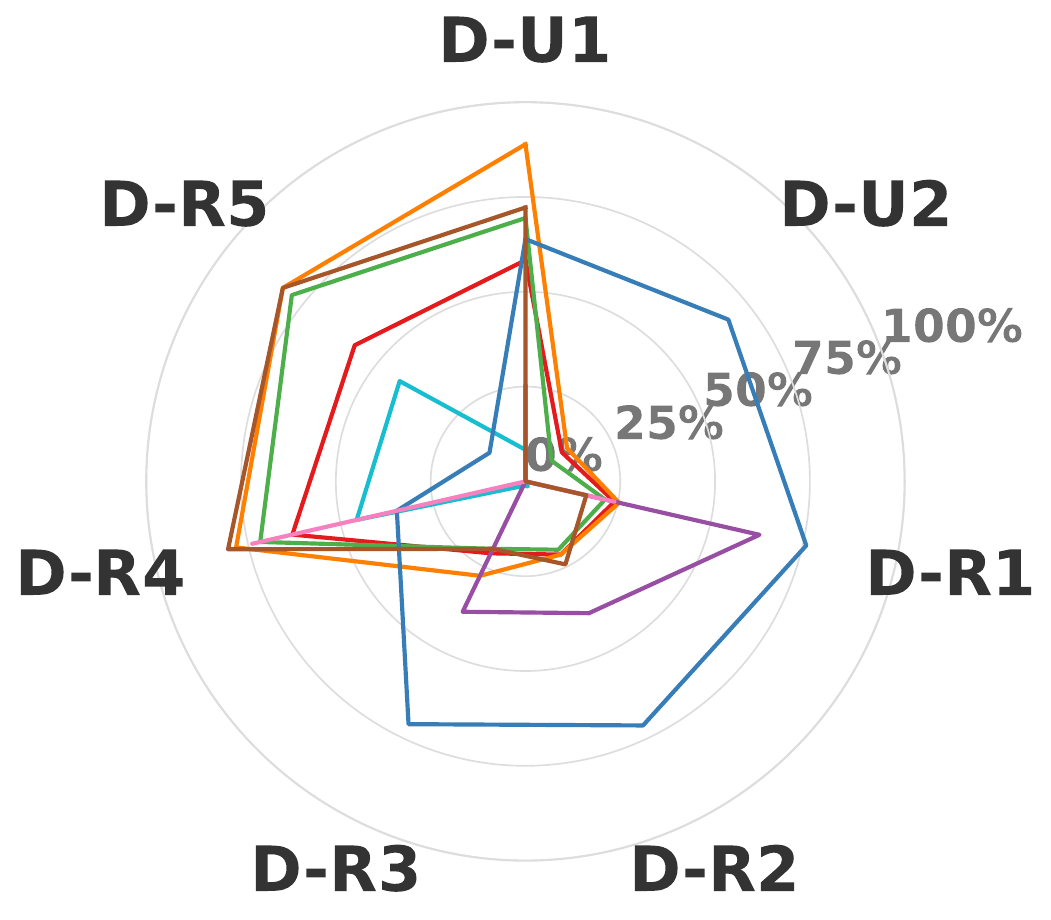}
          \includegraphics[width=0.49\textwidth]{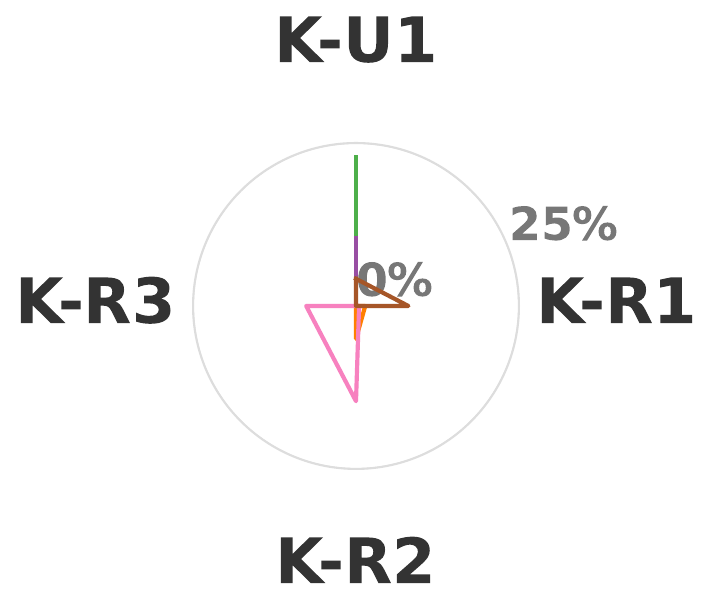}
          \caption{Gemini-2.0}
        \end{subfigure}
        \hfill
        \begin{subfigure}[b]{0.32\textwidth}
          \includegraphics[width=0.49\textwidth]{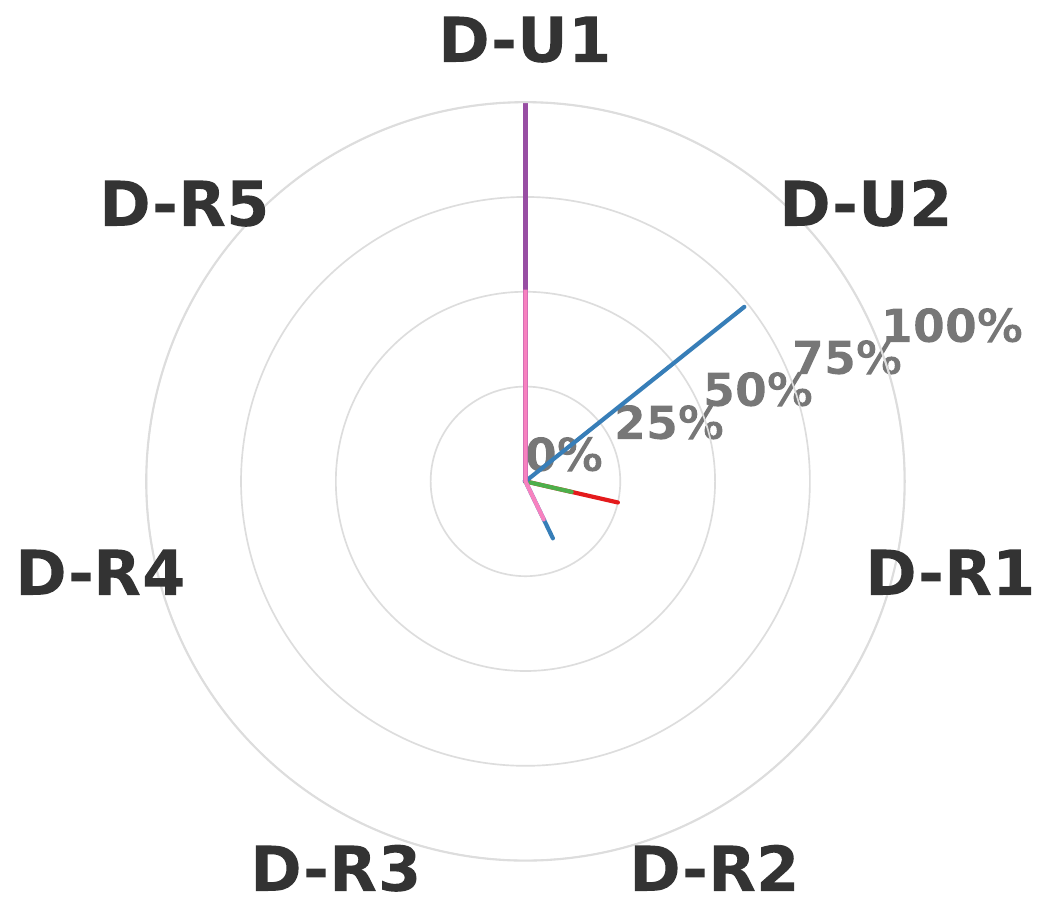}
          \includegraphics[width=0.49\textwidth]{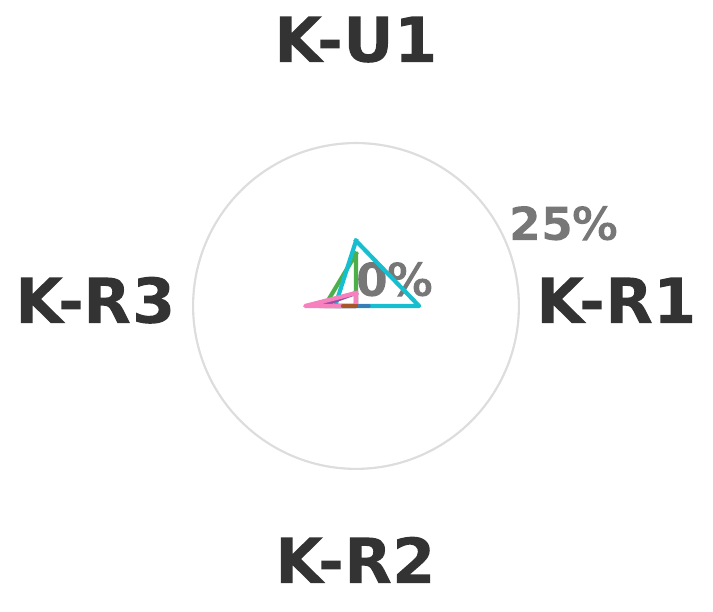}
          \caption{Gemini-2.5}
        \end{subfigure}
        \subcaption*{Non-Medical : 
\textcolor[HTML]{E41A1C}{C.L.E.A.R}, 
\textcolor[HTML]{FF7F00}{TaT}, 
\textcolor[HTML]{4DAF4A}{TableMaster}, 
\textcolor[HTML]{17BECF}{TIDE}, 
\textcolor[HTML]{377EB8}{E5}, 
\textcolor[HTML]{984EA3}{GraphOTTER}, 
\textcolor[HTML]{F781BF}{H-STAR}, 
\textcolor[HTML]{A65628}{Table-R1}, }
      \end{subfigure}

      \begin{subfigure}[b]{\textwidth}
        \centering
        \begin{subfigure}[b]{0.32\textwidth}
          \includegraphics[width=0.49\textwidth]{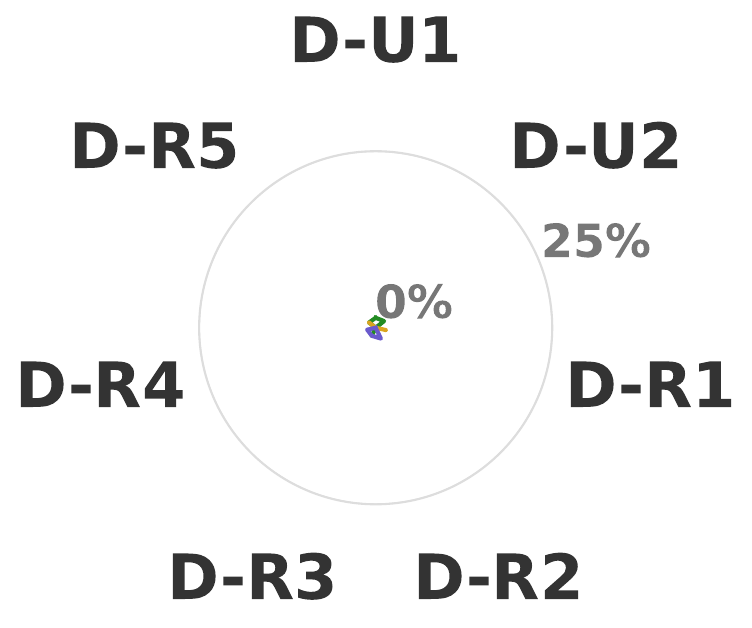}
          \includegraphics[width=0.49\textwidth]{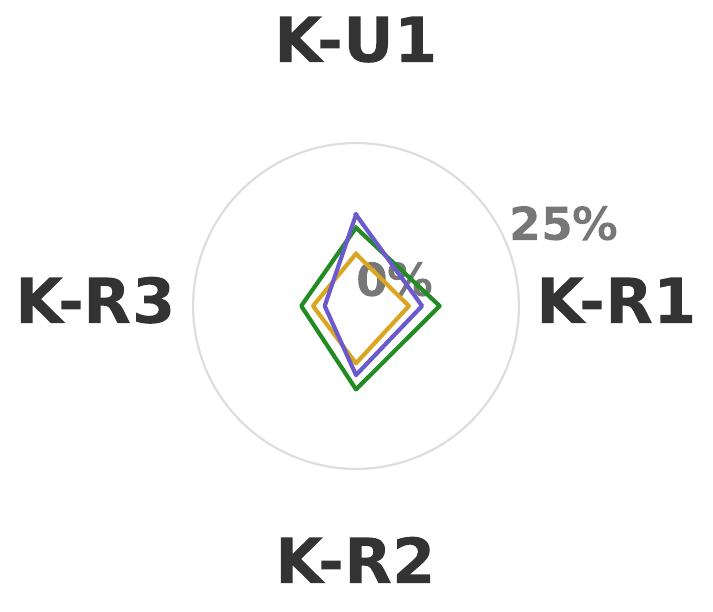}
          \caption{Gemini-1.5}
        \end{subfigure}
        \hfill
        \begin{subfigure}[b]{0.32\textwidth}
          \includegraphics[width=0.49\textwidth]{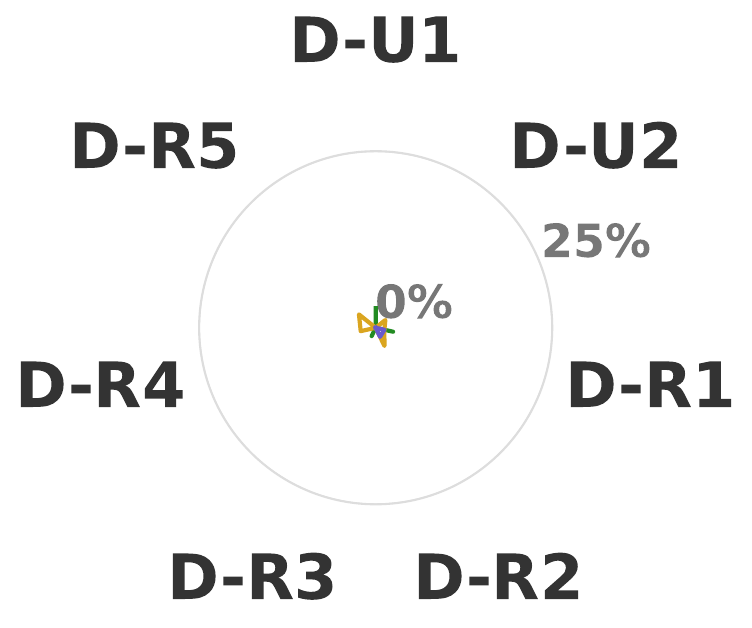}
          \includegraphics[width=0.49\textwidth]{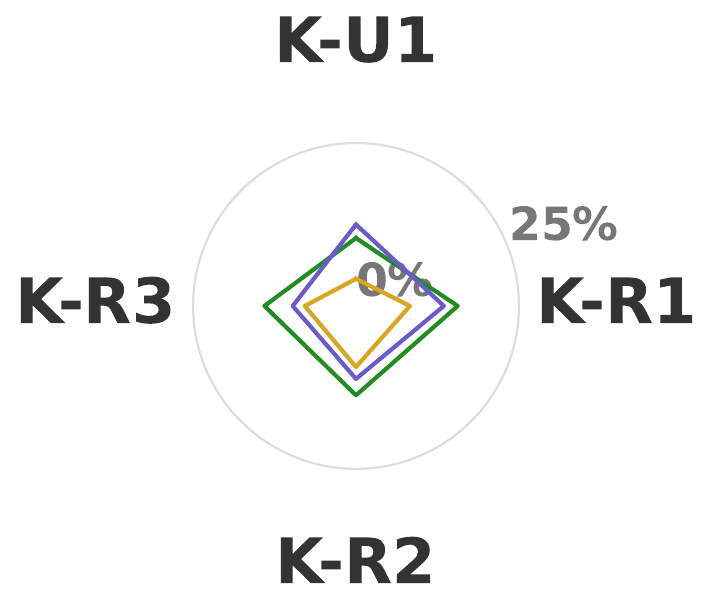}
          \caption{Gemini-2.0}
        \end{subfigure}
        \hfill
        \begin{subfigure}[b]{0.32\textwidth}
          \includegraphics[width=0.49\textwidth]{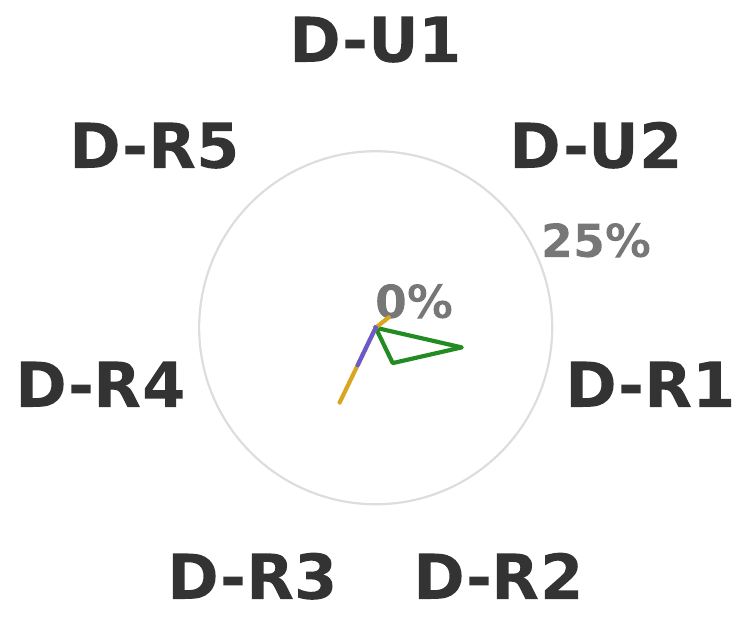}
          \includegraphics[width=0.49\textwidth]{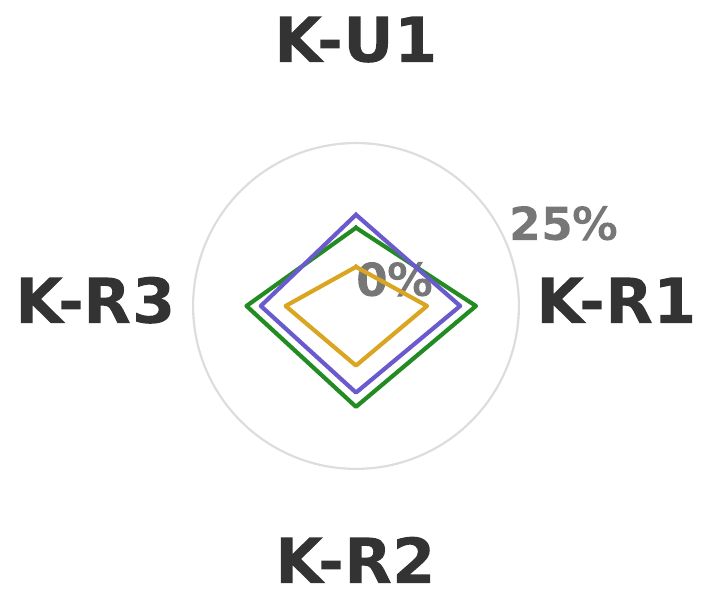}
          \caption{Gemini-2.5}
        \end{subfigure}
        \subcaption*{Medical : \textcolor[HTML]{228B22}{llm4healthcare}, 
    \textcolor[HTML]{DAA520}{DeLLiriuM}, 
    \textcolor[HTML]{6A5ACD}{EnsembleLLM}.
    }
      \end{subfigure}
    
\caption{Comparison of relative gains for 11 SOTA methods across tasks. 
Relative gain is defined as the percentage of improvement each method achieves toward the maximum possible gain for each task, where 0\% indicates no improvement and 100\% represents the upper bound. 
In each subfigure, the left side shows Data-Driven tasks, and the right side shows Knowledge-Driven tasks.}
      \label{fig:gemini_compare}
    \end{figure*}

\begin{table*}[ht]
\begin{tabular}{ccccccccccccc}
\hline
\multirow{3}{*}{Models}    & \multirow{3}{*}{Methods} & \multicolumn{7}{c}{Data-Driven}                                                                     & \multicolumn{4}{c}{Knowledge-Driven}                        \\ \cline{3-13} 
                           &                          & D-U1         & D-U2         & D-R1        & D-R2        & D-R3        & D-R4         & D-R5         & K-U1        & K-R1          & K-R2          & K-R3          \\ \cline{3-13} 
                           &                          & ACC          & ACC          & ACC         & ACC         & ACC         & ACC          & ACC          & AUC         & AUC           & AUC           & AUC           \\ \hline
\multirow{2}{*}{Gemini 1.5} & EHRMaster           & \textbf{100} & \textbf{100} & \textbf{96} & \textbf{96} & \textbf{94} & \textbf{100} & \textbf{100} & \textbf{89} & \textbf{62.3} & 54            & \textbf{54.7} \\
                           &  previous SOTA               & 76           & 79           & 80          & 78          & 73          & 85           & 93           & 57          & 61.3          & \textbf{56.4} & 54.2          \\ \hline
\multirow{2}{*}{Gemini 2.0} & EHRMaster           & \textbf{98}  & \textbf{100} & \textbf{91} & \textbf{81} & \textbf{93} & 80           & 87           & \textbf{67} & \textbf{65.3} & \textbf{64.2} & 56.2          \\
                           &  previous SOTA               & 96           & 82           & 81          & 80          & 78          & \textbf{90}  & \textbf{94}  & 63          & 64.3          & 62.2          & \textbf{58.4} \\ \hline
\multirow{2}{*}{Gemini 2.5} & EHRMaster           & \textbf{100} & \textbf{100} & \textbf{97} & \textbf{95} & \textbf{97} & \textbf{100} & \textbf{100} & \textbf{60} & 59.3          & 55.1          & \textbf{69.2} \\
                           &  previous SOTA               & \textbf{100} & 89           & 94          & 85          & 85          & 100          & 100          & 57          & \textbf{66.3} & \textbf{61.2} & 58.4          \\ \hline
\end{tabular}
\caption{Performance of EHRMaster compared to LLM-based enhancement methods on benchmark.}
\label{tab:EHRMaster_results}
\end{table*}

\subsection{Evaluating LLM-Based Enhancement Methods} \label{sec:sota}
We reproduce and evaluate 11 representative LLM-based enhancement methods designed to enhance performance on structured data tasks. Among them, eight methods were originally developed for non-medical tasks—C.L.E.A.R.~\cite{deng2024enhancing}, TaT~\cite{sun2025table}, TableMaster~\cite{cao2025tablemaster}, TIDE~\cite{yang2025triples}, E\textsuperscript{5}~\cite{zhang2024e5}, GraphOTTER~\cite{li2024graphotter}, H-STAR~\cite{abhyankar2024h}, and Table-R1~\cite{yang2025table}—while the remaining three are specifically designed for medical applications: LLM4Healthcare~\cite{zhu2024prompting}, DeLLiriuM~\cite{contreras2024dellirium}, and EnsembleLLM~\cite{hu2024power}.

As shown in the 6 subfigures of Figure~\ref{fig:gemini_compare}, these methods exhibit clear specialization based on task scenarios. Non-medical methods generally achieve greater gains on Data-Driven tasks—such as field filtering or numeric reasoning—but show limited improvements on Knowledge-Driven tasks that require clinical expertise. In contrast, medical methods perform better on Knowledge-Driven tasks—such as mortality prediction or treatment planning—yet struggle to generalize to Data-Driven scenarios. This divergence highlights a key limitation of existing enhancement approaches: none offer consistent improvements across the full spectrum of structured EHR tasks. These results underscore the need for unified solutions capable of both logical reasoning over structured tables and integration of clinical domain knowledge.

\subsection{EHRMaster Analysis} \label{sec:EHRMaster_analysis}
Table~\ref{tab:EHRMaster_results} reports the performance of our proposed EHRMaster method compared to the SOTA LLM-based enhancement baselines across all benchmark tasks. On Data-Driven tasks, EHRMaster consistently achieves perfect scores across Gemini models, demonstrating strong capabilities in structured reasoning and arithmetic operations. Significant gains appear on arithmetic-heavy tasks such as D-R4 and D-R5, where EHRMaster achieves 100\% accuracy in most cases. For Knowledge-Driven tasks, EHRMaster delivers noticeable improvements on challenging tasks (e.g., K-R2 and K-R3), although gains vary across models and tasks. These results highlight EHRMaster’s effectiveness in structured EHR reasoning, especially in Data-Driven scenarios, while also offering competitive performance on clinically complex tasks.

\section{Conclusion} \label{sec:conclusion}
In summary, we present EHRStruct, a benchmark for evaluating LLMs on structured EHR tasks, enabling systematic comparison across diverse scenarios. EHRStruct addresses the key challenges in applying LLMs to structured EHR data by providing a novel dataset, defining clear task specifications, and establishing a standardized evaluation framework. First, we construct a novel dataset containing 2,200 structured EHR tables, generated synthetically using Synthea and extracted from the real-world eICU database. This dataset spans a wide range of clinical scenarios to ensure robust evaluation. Second, we define 11 tasks across 6 categories, organized into Data-Driven and Knowledge-Driven scenarios, covering a broad spectrum of structured EHR applications, from data extraction to clinical decision support. Finally, we establish a standardized evaluation process, testing LLMs under various settings such as zero-shot, few-shot, and fine-tuning, and analyzing the impact of different input formats. This framework enables consistent and comprehensive assessment of LLM performance on structured EHR tasks.  We also reproduce 11 state-of-the-art (SOTA) LLM-based enhancement methods for structured data processing and evaluate them using our benchmark. Through this evaluation, we identified the limitations of current methods and proposed EHRMaster, a novel code-augmented framework, to address these challenges and achieve superior performance. 

In future work, we plan to extend EHRStruct to evaluate iterative and adaptive treatment-planning processes that incorporate real-time patient responses such as drug reactions and physiological indicators, better reflecting the dynamic nature of clinical decision-making.

\bibliography{aaai2026}

\clearpage
\appendix

\setcounter{secnumdepth}{2} 
\renewcommand{\thesubsection}{\Alph{section}.\arabic{subsection}}

\begin{center}
    \LARGE \textbf{Appendix}
\end{center}

\section{Task} 

\subsection{Task Classification Criteria}\label{sec:detailed_task_classification}

We organize benchmark tasks using a hierarchical classification scheme comprising 4 levels:  
Task Scenario, Task Level, Task Category, and Task ID, listed from high to low granularity.

\begin{itemize}
    \item Task Scenario: Indicates the type of information dependency. It includes:
    \begin{itemize}
        \item Data-Driven: Tasks that can be solved using structured EHR data alone, without relying on domain-specific medical knowledge.
        \item Knowledge-Driven: Tasks requiring external medical knowledge or clinical interpretation beyond raw EHR data.
    \end{itemize}

    \item Task Level: Represents the reasoning complexity of the task:
    \begin{itemize}
        \item Understanding: Tasks that retrieve or extract existing information directly from the input, without combining multiple data elements or requiring inference.
        \item Reasoning: Tasks that involve computation, aggregation, inference, or medical decision-making based on cross-field or cross-time reasoning.
    \end{itemize}

    \item Task Category: Denotes the functional type of the task, such as \textit{Information retrieval}, \textit{Data aggregation}, or \textit{Diagnostic assessment}.  
    These categories are extensible to accommodate additional task types in future benchmarks.

    \item Task ID: Specifies the unique identifier of a benchmark instance (e.g., D-R2, K-U1), which reflects its scenario, level, and category.
\end{itemize}

\subsection{Task Definitions}
\label{sec:detailed_task_description}

We define 6 task categories in our benchmark to evaluate different capabilities of large language models over structured EHR data.  
Each category corresponds to a specific functional objective commonly encountered in clinical workflows, and is grounded in either data-centric retrieval or knowledge-informed reasoning.

\begin{enumerate}[label=\roman*)]

\item \textit{Information Retrieval}:
This category supports clinicians in locating specific values from structured EHR records, such as lab test results or demographic attributes.
Large language models have recently been explored for performing retrieval over structured inputs.
Our benchmark includes two tasks (D-U1, D-U2) representing this category in realistic data-driven scenarios, defined by logical comparison direction (e.g., greater, less) for field-based filtering.

\item \textit{Data Aggregation}:
This category involves combining multiple EHR entries into aggregate statistics, such as counts, means, or sums.
Traditionally performed manually or via scripting, recent work shows that LLMs can automate such aggregation operations over tabular inputs.
Our benchmark includes three tasks (D-R1, D-R2, D-R3) that cover common arithmetic operations—Count, Average, and Sum.

\item \textit{Arithmetic Computation}:
In many clinical tasks, structured values must be compared, subtracted, or combined—such as tracking weight changes, computing risk scores, or identifying abnormal fluctuations.
This category evaluates a model’s ability to perform simple arithmetic over structured features, without explicit numerical formulas provided.
Our benchmark includes two tasks (D-R4, D-R5) involving arithmetic reasoning operations such as subtraction and addition across related numeric fields.

\item \textit{Clinical Identification}:
Identifying clinical concepts, such as disease mentions or structured diagnosis codes, is foundational for building downstream applications.
Despite the prevalence of structured data in EHRs, clinical concept recognition remains underexplored in LLM benchmarks.
Our benchmark includes task K-U1 to assess whether models can identify structured clinical conditions like SNOMED-CT diagnoses, representing the knowledge-level identification type.

\item \textit{Diagnostic Assessment}:
This category evaluates whether models can reason over clinical variables to make diagnostic inferences, such as disease likelihood or mortality risk.
It reflects real-world settings where multiple lab values, symptoms, and comorbidities must be synthesized to support clinical decisions.
Our benchmark includes two tasks (K-R1, K-R2) focusing on mortality prediction and disease classification based on structured inputs, corresponding to reasoning-level diagnosis and risk assessment.

\item \textit{Treatment Planning}:
Given a set of diagnosed conditions or clinical history, determining an appropriate treatment plan is a high-level clinical reasoning task.
This category evaluates a model’s ability to recommend treatment strategies, such as medication selection, based on structured diagnoses.
Our benchmark includes task K-R3, where models must select suitable drug classes, representing the reasoning-level treatment planning type.

\end{enumerate}

Together, these 6 categories span a diverse range of clinically relevant task types, from factual retrieval to complex reasoning.  
They provide a comprehensive testbed for evaluating both the information extraction and decision-making capabilities of LLMs over structured healthcare data.

\subsection{Task Instructions}
\label{sec:all_instructions}

We provide the complete set of prompt instructions used in our benchmark, organized by task ID.  
Each instruction follows a standardized format that includes the input table, task description, and output requirements.  
These instructions are used both for prompting large language models and for evaluating their consistency across diverse clinical scenarios.  
Appendix shows a single instruction example for each task, while the actual benchmark contains multiple samples, each paired with its own context-specific instruction.

\onecolumn

\newtcolorbox{instructionbox}[1][]{
  colback=green!5,
  colframe=green!50!black,
  fonttitle=\bfseries,
  title=#1,
  boxrule=0.5pt,
  arc=2mm,
  outer arc=2mm,
  left=1mm,
  right=1mm,
  top=1mm,
  bottom=1mm,
  breakable,
}

\textbf{Synthetic Dataset.}

\begin{instructionbox}[D-U1 Instruction (Synthea)]
You are provided with a structured table containing patient information.

Table:
\begin{tabular}{l|l|l|l}
ID & RACE & GENDER & INCOME \\
\hline
001 & White & Male & 45000 \\
002 & Black & Female & 52000 \\
003 & White & Female & 28000 \\
004 & White & Male & 36000 \\
\end{tabular}

Task:

Find all patients who are white and have an income greater than 30000.

Instructions:

Return only the list of corresponding IDs, separated by commas. 

Use `NULL` if no patients meet the criteria.

Output format:

D-U1: ID1,ID2,...

Important:

Do NOT explain. Only return the line above.
\end{instructionbox}

\begin{instructionbox}[D-U2 Instruction (Synthea)]
You are provided with a structured table containing patient information.

Table:
\begin{tabular}{l|l|l|l}
ID & RACE & GENDER & INCOME \\
\hline
005 & Asian & Female & 18000 \\
006 & White & Male & 24000 \\
007 & Black & Female & 15000 \\
008 & White & Female & 22000 \\
\end{tabular}

Task:

Find all patients who are female and have an income less than 20000.

Instructions:

Return only the list of corresponding IDs, separated by commas. 

Use `NULL` if no patients meet the criteria.

Output format:

D-U2: ID1,ID2,...

Important:

Do NOT explain. Only return the line above.
\end{instructionbox}

\begin{instructionbox}[D-R1 Instruction (Synthea)]
You are given a patient table. For the patient ID = 002, answer the following about their pain severity scores:

Table:
\begin{tabular}{l|l|l|l|l}
PATIENT & DESCRIPTION & VALUE & UNITS & TYPE \\
\hline
001 & Body Weight & 71.4 & kg & numeric \\
002 & Pain severity & 2 & score & numeric \\
002 & Pain severity & 4 & score & numeric \\
002 & Heart rate & 88 & min & numeric \\
003 & Pain severity & 1 & score & numeric \\
002 & Respiratory Rate & 16 & min & numeric \\
001 & Pain severity & 3 & score & numeric \\
\end{tabular}

Task:

How many pain severity scores are recorded?

Instructions:

Do NOT explain.

Output format:

D-R1: [number]

Important:

Only return the number as shown above.
\end{instructionbox}

\begin{instructionbox}[D-R2 Instruction (Synthea)]
You are given a patient table. For the patient ID = 003, answer the following about their pain severity scores:

Table:
\begin{tabular}{l|l|l|l|l}
PATIENT & DESCRIPTION & VALUE & UNITS & TYPE \\
\hline
001 & Pain severity & 5 & score & numeric \\
003 & Pain severity & 2 & score & numeric \\
002 & Heart rate & 91 & min & numeric \\
003 & Pain severity & 3 & score & numeric \\
003 & Respiratory Rate & 17 & min & numeric \\
002 & Pain severity & 1 & score & numeric \\
003 & Pain severity & 4 & score & numeric \\
\end{tabular}

Task:

What is the average pain severity score? (rounded to 1 decimal place)

Instructions:

Do NOT explain.

Output format:

D-R2: [number]

Important:

Only return the number as shown above.
\end{instructionbox}

\begin{instructionbox}[D-R3 Instruction (Synthea)]
You are given a patient table. For the patient ID = 001, answer the following about their pain severity scores:

Table:
\begin{tabular}{l|l|l|l|l}
PATIENT & DESCRIPTION & VALUE & UNITS & TYPE \\
\hline
001 & Pain severity & 2 & score & numeric \\
002 & Respiratory Rate & 14 & min & numeric \\
003 & Pain severity & 1 & score & numeric \\
001 & Pain severity & 5 & score & numeric \\
002 & Pain severity & 3 & score & numeric \\
001 & Heart rate & 99 & min & numeric \\
001 & Pain severity & 4 & score & numeric \\
\end{tabular}

Task:

What is the total pain severity score?

Instructions:

Do NOT explain.

Output format:

D-R3: [number]

Important:

Only return the number as shown above.
\end{instructionbox}

\begin{instructionbox}[D-R4 Instruction (Synthea)]
Table:
\begin{tabular}{l|l|l|r|r}
ID & RACE & GENDER & HEALTHCARE & INCOME \\
\hline
001 & white & M & 20350.20 & 34899.50 \\
002 & asian & F & 80000.00 & 62342.75 \\
003 & white & F & 145000.35 & 179235.89 \\
004 & black & M & 48000.00 & 39880.00 \\
005 & white & F & 98200.00 & 105000.00 \\
\end{tabular}

Task:

For patient ID = 003, how much more is their income than their healthcare expenses? (round to 2 decimals)

Instructions:

Return exactly the number.  

For example: -13725.74  

Do NOT explain. Do NOT add extra text. Only return the numbers.

Output format:

D-R4: [number]
\end{instructionbox}

\begin{instructionbox}[D-R5 Instruction (Synthea)]
Table:
\begin{tabular}{l|l|l|r|r}
ID & RACE & GENDER & HEALTHCARE & INCOME \\
\hline
001 & white & F & 98200.00 & 105000.00 \\
002 & asian & M & 45000.50 & 39200.00 \\
003 & white & F & 67890.25 & 82450.75 \\
004 & black & F & 120000.00 & 139500.00 \\
005 & asian & F & 78900.00 & 89400.00 \\
\end{tabular}

Task:

For patient ID = 003, what is the total amount of this patient's income and healthcare expenses combined? (round to 2 decimals)

Instructions:

Return exactly the number.  

For example: 122145.35  

Do NOT explain. Do NOT add extra text. Only return the numbers.

Output format:
D-R5: [number]
\end{instructionbox}

\begin{instructionbox}[K-U1 Instruction (Synthea)]
Table:
\begin{tabular}{l|l|l|l}
START & STOP & SYSTEM & CODE \\
\hline
2005/04/12 & 2006/11/20 & SNOMED-CT & 230690007 \\
2007/08/03 & 2010/02/17 & SNOMED-CT & 11687002 \\
2011/05/21 & 2014/09/30 & SNOMED-CT & 44054006 \\
2015/01/13 & 2018/08/07 & SNOMED-CT & 15777000 \\
2019/03/15 & 2020/10/19 & SNOMED-CT & 195967001 \\
\end{tabular}

Task:
The 'CODE' column contains SNOMED-CT codes representing clinical conditions diagnosed for this patient.  

Has this patient ever shown symptoms related to diabetes, based on these clinical codes?

Instructions:
Return in the format:  

Diabetes: 0 or 1

Do NOT explain. Only return the prediction line.
\end{instructionbox}

\begin{instructionbox}[K-R1 Instruction (Synthea)]
The following table contains structured medical observation data for one patient.  

Table:
\begin{tabular}{l|l|l|l}
DATE & DESCRIPTION & VALUE & UNITS \\
\hline
2001-02-15 & Body temperature & 37.6 & Cel \\
2001-02-15 & Pain severity & 2 & score \\
2001-02-15 & Body Mass Index & 31.2 & kg/m2 \\
2001-02-15 & Heart rate & 84 & min \\
2001-02-15 & Tobacco status & Current every day smoker &  \\
2001-02-15 & Housing status & Homeless &  \\
2001-02-15 & Employment status & Unemployed &  \\
2001-02-15 & Do you feel hopeless & Yes &  \\
2001-02-15 & In the past week, have you had trouble sleeping & Yes &  \\
\end{tabular}

Task:  

Based on the table, do you think this patient is likely to die during the simulation?  

Instructions:  

Return in the format:  

Death: 0 or 1

Do NOT explain.

Only return the prediction line.
\end{instructionbox}

\begin{instructionbox}[K-R2 Instruction (Synthea)]
Table:

\begin{tabular}{l|l|l|l}
DATE & DESCRIPTION & VALUE & UNITS \\
\hline
2019-08-25 & Body Height & 168.4 & cm \\
2019-08-25 & Pain severity & 4 & score \\
2019-08-25 & Body Weight & 71.8 & kg \\
2019-08-25 & Body mass index & 25.2 & kg/m2 \\
2019-08-25 & Body mass percentage & 30 & \% \\
2019-08-25 & Diastolic Blood Pressure & 85 & mm[Hg] \\
2019-08-25 & Systolic Blood Pressure & 118 & mm[Hg] \\
2019-08-25 & Heart rate & 91 & min \\
2019-08-25 & Respiratory Rate & 16 & min \\
2019-08-25 & Erythrocytes & 4.9 & 10*6/uL \\
2019-08-25 & Hematocrit & 40.1 & \% \\
2019-08-25 & MCV & 86.4 & fL \\
2019-08-25 & MCH & 29.5 & pg \\
2019-08-25 & MCHC & 34.2 & g/dL \\
2019-08-25 & Platelets [\#] & 212.3 & 10*3/uL \\
2019-08-25 & Platelet distribution width & 145.6 & fL \\
\end{tabular}

Task:  

Does this patient likely have 'Acute viral pharyngitis (disorder)'?  

Instructions:  

Return in the format:  

Disorder: 0 or 1 

Do NOT explain.  

Only return the prediction line.  
\end{instructionbox}

\begin{instructionbox}[K-R3 Instruction (Synthea)]
Table:

\begin{tabular}{l|l|l|l}
DATE & DESCRIPTION & VALUE & UNITS \\
\hline
2018-07-12 & Body temperature & 39.1 & Cel \\
2018-07-12 & Pain severity & 5 & score \\
2018-07-12 & Diastolic Blood Pressure & 75 & mm[Hg] \\
2018-07-12 & Systolic Blood Pressure & 112 & mm[Hg] \\
2018-07-12 & Heart rate & 95 & min \\
2018-07-12 & Respiratory Rate & 15 & min \\
2018-07-12 & Tobacco status & Never smoked &  \\
2018-07-12 & Employment status & Employed full-time &  \\
2018-07-12 & Generalized discomfort & 6 & score \\
2018-07-12 & Cough frequency & High &  \\
2018-07-12 & Throat redness & Present &  \\
\end{tabular}

Task:  

Based on the clinical history, would penicillin be an appropriate treatment option for this patient?  

Instructions:  

Return in the format:  

Recommend: 0 or 1  

Do NOT explain.  

Only return the prediction line.  
\end{instructionbox}

\newtcolorbox{instructionbox_eicu}[1][]{
  colback=orange!5,
  colframe=orange!50!black,
  fonttitle=\bfseries,
  title=#1,
  boxrule=0.5pt,
  arc=2mm,
  outer arc=2mm,
  left=1mm,
  right=1mm,
  top=1mm,
  bottom=1mm,
  breakable,
}

\textbf{eICU Dataset.}

\begin{instructionbox_eicu}[D-U1 Instruction (eICU)]
Table:

\begin{tabular}{l|l|l|l|l}
patientunitstayid & gender & age & ethnicity & hospitaldischargestatus \\
\hline
001 & Female & 86 & african american & expired \\
002 & Female & 72 & caucasian & alive \\
003 & Male & 67 & african american & expired \\
004 & Female & 90 & african american & expired \\
005 & Female & 59 & african american & alive \\
006 & Female & 73 & african american & expired \\
007 & Male & 84 & caucasian & expired \\
008 & Female & 61 & african american & alive \\
009 & Male & 70 & african american & alive \\
010 & Female & $>$ 89 & african american & expired \\
\end{tabular}

Task:  

Find all American female patients older than 60 years who died.  

Instructions:  

Return the result in this format:  

D-U1: ID1,ID2,...  

Do NOT explain. 

Only return the line.  
\end{instructionbox_eicu}

\begin{instructionbox_eicu}[D-U2 Instruction (eICU)]
Table:

\begin{tabular}{l|l|l|l|l}
patientunitstayid & gender & age & ethnicity & hospitaldischargestatus \\
\hline
001 & Male & 72 & caucasian & alive \\
002 & Female & 65 & african american & expired \\
003 & Male & 89 & caucasian & alive \\
004 & Male & > 89 & caucasian & expired \\
005 & Male & 47 & caucasian & alive \\
006 & Female & 84 & caucasian & alive \\
007 & Male & 73 & caucasian & expired \\
008 & Male & 60 & caucasian & alive \\
009 & Female & 90 & caucasian & alive \\
010 & Male & 67 & african american & alive \\
\end{tabular}

Task:  

Find all Caucasian male patients younger than 90 years who were discharged alive.  

Instructions:  

Return the result in this format:  

D-U2: ID1,ID2,...  

Do NOT explain. 

Only return the line.  
\end{instructionbox_eicu}

\begin{instructionbox_eicu}[D-R1 Instruction (eICU)]
You are given a patient table. For the patient ID = 002, answer the following about their temperature values:  

Table:

\begin{tabular}{l|l|l|l}
PATIENT & DESCRIPTION & UNITS & VALUE \\
\hline
001 & Temperature & TEMP ORAL & 36.6 \\
002 & Temperature & TEMP TYMPANIC & 37.4 \\
002 & O2 Saturation & O2 Sat & 94 \\
003 & Temperature & TEMP ORAL & 38.1 \\
002 & Temperature & TEMP ORAL & 36.9 \\
002 & Respiratory Rate & Resp & 18 \\
001 & Blood Pressure & BP & 120/80 \\
002 & Temperature & TEMP ORAL & 37.0 \\
003 & Pain Assessment & WDL &  \\
002 & Temperature & TEMP TYMPANIC & 36.8 \\
\end{tabular}

Task:  

How many temperature records are present?  

Instructions:  

Do NOT explain.  

Return the number only.  
\end{instructionbox_eicu}

\begin{instructionbox_eicu}[D-R2 Instruction (eICU)]
You are given a patient table. For the patient ID = 001, answer the following about their temperature values:  

Table:

\begin{tabular}{l|l|l|l}
PATIENT & DESCRIPTION & UNITS & VALUE \\
\hline
001 & Temperature & TEMP ORAL & 36.5 \\
002 & Temperature & TEMP TYMPANIC & 37.2 \\
001 & Temperature & TEMP TYMPANIC & 36.8 \\
003 & O2 Saturation & O2 Sat & 97 \\
001 & Respiratory Rate & Resp & 16 \\
001 & Temperature & TEMP ORAL & 36.9 \\
001 & Temperature & TEMP TYMPANIC & 37.0 \\
002 & Pain Assessment & WDL &  \\
003 & Temperature & TEMP ORAL & 38.3 \\
\end{tabular}

Task:  

What is the average temperature value? (in Celsius, rounded to 1 decimal place)  

Instructions:  

Do NOT explain.  

Return the number only.  
\end{instructionbox_eicu}

\begin{instructionbox_eicu}[D-R3 Instruction (eICU)]
You are given a patient table. For the patient ID = 003, answer the following about their temperature values:  

Table:

\begin{tabular}{l|l|l|l}
PATIENT & DESCRIPTION & UNITS & VALUE \\
\hline
003 & Temperature & TEMP ORAL & 36.4 \\
001 & Respiratory Rate & Resp & 15 \\
003 & Temperature & TEMP TYMPANIC & 37.2 \\
002 & Pain Assessment & WDL &  \\
003 & Temperature & TEMP ORAL & 36.9 \\
003 & O2 Saturation & O2 Sat & 98 \\
003 & Temperature & TEMP TYMPANIC & 37.1 \\
001 & Temperature & TEMP ORAL & 36.5 \\
\end{tabular}

Task:  

What is the total temperature value? (sum, in Celsius, rounded to 1 decimal place)  

Instructions:  

Do NOT explain.  

Return the number only.  
\end{instructionbox_eicu}

\begin{instructionbox_eicu}[D-R4 Instruction (eICU)]
Table:

\begin{tabular}{l|l|l|l|l|l|l|l}
age & tax & gender & patientunitstayid & admissionweight & unitvisitnumber & cost & dischargeweight \\
\hline
55 & 1450.2 & Female & 001 & 70.5 & 1 & 8234.6 & 68.4 \\
72 & 2130.5 & Male & 002 & 85.0 & 1 & 16789.2 & 86.3 \\
64 & 589.3 & Female & 003 & 91.2 & 1 & 14350.0 & 88.1 \\
60 & 984.1 & Female & 004 & 64.0 & 1 & 7300.2 & 64.0 \\
45 & 2050.7 & Male & 005 & 78.3 & 1 & 19240.8 & 82.9 \\
\end{tabular}

Task:  

how much did their weight change during the hospital stay? (rounded to 1 decimal place)  

Instructions:  

Return exactly number  

For example: 1.5  

Do NOT explain.  

Do NOT add extra text.  

Only return the numbers.  
\end{instructionbox_eicu}

\begin{instructionbox_eicu}[D-R5 Instruction (eICU)]
Table:

\begin{tabular}{l|l|l|l|l|l|l|l}
age & tax & gender & patientunitstayid & admissionweight & unitvisitnumber & cost & dischargeweight \\
\hline
76 & 1012.3 & Female & 001 & 65.2 & 1 & 11400.7 & 66.1 \\
58 & 875.0 & Male & 002 & 82.0 & 1 & 9800.3 & 84.0 \\
58 & 875.0 & Male & 002 & 82.0 & 2 & 10200.5 & 83.5 \\
43 & 1903.2 & Female & 003 & 70.3 & 1 & 7600.1 & 69.0 \\
67 & 1520.7 & Male & 004 & 90.4 & 1 & 13400.0 & 91.3 \\
\end{tabular}

Task:  

What is the total amount this patient was charged for their stay? (rounded to 1 decimal place)  

Instructions:  

Do NOT explain.  

Only return the number.  
\end{instructionbox_eicu}

\begin{instructionbox_eicu}[K-U1 Instruction (eICU)]
Table:

\begin{tabular}{l|l|l|l}
START & STOP & SYSTEM & CODE \\
\hline
2018-05-01 & 2018-06-14 & SNOMED-CT & 386661006 \\
2018-07-20 & 2018-09-03 & SNOMED-CT & 161615003 \\
2019-02-15 & 2019-04-20 & SNOMED-CT & 128599005 \\
2019-11-05 & 2020-01-17 & SNOMED-CT & 230572002 \\
2020-07-10 & 2020-09-25 & SNOMED-CT & 233604007 \\
\end{tabular}

Task:  

The 'CODE' column contains SNOMED-CT concepts recorded for this patient during ICU visits.  

Has this patient ever been diagnosed with asthma (disorder) based on these SNOMED-CT codes?  

Instructions:  

Return in the format:  

Asthma: <0 or 1>  

Do NOT explain.  

Only return the prediction line.  
\end{instructionbox_eicu}

\begin{instructionbox_eicu}[K-R2 Instruction (eICU)]
Table:

\begin{tabular}{l|l|l|l|l|l|l}
labid & patientunitstayid & labresultoffset & labtypeid & labname & labresult & labresulttext \\
\hline
51480001 & 100002 & -130 & 3 & WBC x 1000 & 38.2 & 38.2 \\
51480002 & 100002 & -130 & 3 & Hgb & 7.5 & 7.5 \\
51480003 & 100002 & -130 & 3 & platelets x 1000 & 520.0 & 520 \\
51480004 & 100002 & -130 & 1 & creatinine & 3.6 & 3.6 \\
51480005 & 100002 & -130 & 1 & bicarbonate & 14.0 & 14.0 \\
51480006 & 100002 & -130 & 1 & troponin - I & 0.08 & 0.08 \\
51480007 & 100002 & -130 & 1 & BNP & 1980.0 & 1980 \\
51480008 & 100002 & -130 & 1 & sodium & 133.0 & 133 \\
\end{tabular}

Task:  

Based on the above information, please predict whether the patient died in hospital.  

Instructions:  

Only output one word: 'Alive' or 'Expired'.  

Do not explain.  
\end{instructionbox_eicu}

\begin{instructionbox_eicu}[K-R2 Instruction (eICU)]
Table:

\begin{tabular}{l|l|l|l|l|l|l}
labid & patientunitstayid & labresultoffset & labtypeid & labname & labresult & labresulttext \\
\hline
51483001 & 100004 & -100 & 3 & WBC x 1000 & 30.2 & 30.2 \\
51483002 & 100004 & -100 & 3 & Hgb & 8.6 & 8.6 \\
51483003 & 100004 & -100 & 3 & platelets x 1000 & 530.0 & 530 \\
51483004 & 100004 & -100 & 1 & creatinine & 2.9 & 2.9 \\
51483005 & 100004 & -100 & 1 & glucose & 211.0 & 211 \\
51483006 & 100004 & -100 & 1 & BNP & 1650.0 & 1650 \\
51483007 & 100004 & -100 & 1 & troponin - I & 0.06 & 0.06 \\
51483008 & 100004 & -100 & 1 & sodium & 134.0 & 134 \\
51483009 & 100004 & -100 & 1 & anion gap & 20.0 & 20 \\
51483010 & 100004 & -100 & 1 & bicarbonate & 15.0 & 15 \\
\end{tabular}

Task:  

Based on the patient's record below, please predict whether the patient has each of 10 predefined diseases.  

Instructions:  

Respond with exactly 10 values, each being 0 or 1, corresponding to the diseases above in order.  

Only output these 10 values, no explanation.  
\end{instructionbox_eicu}

\begin{instructionbox_eicu}[K-R3 Instruction (eICU)]
Table:

\begin{tabular}{l|l}
ICD9\_CODE & LONG\_TITLE \\
\hline
1985 & Secondary malignant neoplasm of bone and bone marrow \\
25000 & Diabetes mellitus without mention of complication \\
60001 & Hypertrophy of prostate with urinary obstruction \\
7140 & Rheumatoid arthritis \\
99664 & Infection due to indwelling urinary catheter \\
42731 & Atrial fibrillation \\
5070 & Pneumonitis due to inhalation of food or vomitus \\
78039 & Other convulsions \\
\end{tabular}

Task:  

Based on the patient's diagnosed conditions at hospital admission, please determine **for each** of the following drugs whether it should be prescribed. (The list of drugs is predefined and contains 10 items.)  

Instructions:  

Return a list of exactly 10 binary values (0 or 1), each indicating whether the corresponding drug should be prescribed (1 = yes, 0 = no).  

Only output the 10 values, nothing else.  
\end{instructionbox_eicu}
\twocolumn

\section{EHRMaster Framework}
\label{sec:EHRMaster}

\newtcolorbox{instructionbox_stage}[1][]{
  colback=violet!5,
  colframe=violet!60!black,
  fonttitle=\bfseries,
  title=#1,
  boxrule=0.5pt,
  arc=2mm,
  outer arc=2mm,
  left=1mm,
  right=1mm,
  top=1mm,
  bottom=1mm,
  breakable,
}

The EHRMaster framework operates in three stages: Solution Planning, Concept Alignment, and Adaptive Execution.  
This section provides a detailed description of each stage and how they collectively enable structured EHR task solving with LLMs.

\subsection{Solution Planning}
This stage generates a high-level reasoning plan based on the input question.  
The goal is to decompose the problem into a sequence of logical steps that guide the model toward the final answer.  
The output is a natural language plan describing which information to retrieve, compare, compute, or reason about.  
This plan serves as an intermediate representation that bridges the user query and structured data alignment.

To address the diverse nature of clinical tasks, we design 4 distinct instruction formats aligned with each (Scenario, Level) pair:  
Data-Driven Understanding (D-U), Data-Driven Reasoning (D-R), Knowledge-Driven Understanding (K-U), and Knowledge-Driven Reasoning (K-R).  
Each instruction style adapts the planning objective to the complexity and context of the target task—  
from locating factual entries to simulating clinical decision-making with external knowledge. Four instructions are as follows:

\begin{instructionbox_stage}[D-U Planning Instruction]

You are a table question analyzer.  

Given a question, describe its logic as a short natural language summary that includes:  

- The goal: what field or result the user wants  

- The conditions: what constraints the data must satisfy (e.g., gender is female, race is white)  

Use full sentences in English.  

Keep it short and clear.  

\textbf{Example:}  

Question: What is the maximum pain score for patient P001?  

Output: Find the maximum value of pain score for records where patient is P001.  

Now analyze the following task:  

\texttt{\{task\}} \quad \textit{(This is the task description to be analyzed)}  

Output:  

\end{instructionbox_stage}

\begin{instructionbox_stage}[D-R Planning Instruction]

You are a table reasoning planner.  

Given a question, describe the task as a reasoning plan in natural language:  

- Describe what needs to be calculated (e.g., count, average, total)  

- Mention any filtering conditions  

- Mention involved fields and what to do with them  

Use complete English sentences.  

\textbf{Example:}  

Question: What is the total hospitalization cost for patient P002?  

Output: Calculate the sum of hospitalization cost for patient P002.  

Now analyze the following task:  

\texttt{\{task\}} \quad \textit{(This is the task description to be analyzed)}  

Output:  

\end{instructionbox_stage}

\begin{instructionbox_stage}[K-U Planning Instruction]

You are a table reasoning planner with clinical knowledge.  

Your task:  

Given a natural language instruction, rewrite it as a clear and precise logic description in English sentences.  

This logic will be used to generate executable code over structured data.  

When the instruction contains a clinical term (e.g., Prediabetes), you must:  

- Recognize it as a SNOMED-CT concept  

- Explicitly state which SNOMED-CT codes it corresponds to, if known  

- Only match exact conditions (e.g., Prediabetes), and ignore broader or related terms (e.g., diabetes)  

The logic description must:  

- Clearly describe what to compute or check  

- Mention any filtering criteria (e.g., SYSTEM == 'SNOMED-CT')  

- Specify relevant fields (e.g., CODE column)  

- Always include the exact SNOMED-CT codes when available  

\textbf{Example:}  

Question: Has this patient shown signs of \textbf{Prediabetes} based on SNOMED codes?  

Output: Check if the CODE column contains SNOMED-CT code \textbf{714628002}, which corresponds to \textbf{Prediabetes}.  
If such a code is found and SYSTEM is 'SNOMED-CT', then return 'Prediabetes: 1'; otherwise return 'Prediabetes: 0'.  

Now analyze the following task:  

\texttt{\{task\}} \quad \textit{(This is the task description to be analyzed)}  

Only return the logic description in English sentences.  

Do NOT return code or JSON.  

\end{instructionbox_stage}

\begin{instructionbox_stage}[K-R Planning Instruction]

You are a clinical reasoning planner.  

Your task:  

Given a natural language question and a structured table of clinical indicators for a patient, describe the reasoning steps required to reach the decision implied by the question.  

This may include predicting a clinical outcome (e.g., mortality), identifying a likely diagnosis, or recommending a treatment option.  

Guidelines:  

- Identify the clinical objective (e.g., death prediction, disease identification, treatment recommendation)  

- Examine the table and describe which types of evidence should be considered (e.g., vital signs, lab results, symptoms, diagnostic phrases)  

- Specify which fields (e.g., DESCRIPTION, VALUE, UNITS, CODE) are relevant to support the reasoning  

- If the question implies a yes/no decision, describe the clinical logic that would support either outcome — but do not hardcode thresholds unless explicitly given  

- Do NOT return a final answer — only describe the logical process for reaching one  

\textbf{Example 1 (Mortality Prediction):}  

Question: Based on the table, is this patient likely to die during the simulation?  

Output: Review the patient’s vital signs and lab observations for signs of clinical deterioration, such as sustained high fever, low blood pressure, reduced oxygen saturation, or high-risk diagnoses (e.g., sepsis).  
If multiple critical indicators are present, predict 'Death: 1'; otherwise, predict 'Death: 0'.  

\textbf{Example 2 (Disorder Identification):}  

Question: Does this patient likely have 'Acute viral pharyngitis (disorder)'?  

Output: Look for signs such as sore throat, fever, or viral respiratory symptoms in the DESCRIPTION and VALUE fields.  
If the clinical evidence supports a diagnosis of acute viral pharyngitis, predict 'Disorder: 1'; otherwise, predict 'Disorder: 0'.  

\textbf{Example 3 (Treatment Recommendation):}  

Question: Should penicillin be recommended for this patient?  

Output: Identify symptom patterns that suggest bacterial infection (e.g., high fever, pharyngeal inflammation), and also check for signs of allergy or contraindication.  
If clinical indicators support the use of penicillin and no risks are found, predict 'Recommend: 1'; otherwise, predict 'Recommend: 0'.  

Now analyze the following task:  

\texttt{\{task\}} \quad \textit{(This is the task description to be analyzed)}  

Only return the reasoning logic in English sentences.  

Do NOT return code or JSON.  

\end{instructionbox_stage}

\subsection{Concept Alignment}

This stage connects the reasoning plan generated in the Solution Planning phase with the actual structure of the input EHR table.  

The goal is to identify which columns, fields, or data segments in the structured input correspond to the concepts mentioned in the plan.  
This includes mapping abstract entities—such as “blood glucose level” or “admission date”—to their concrete representations in the tabular data.  

To perform this mapping consistently across all task types, we use a unified instruction format described below.  

\begin{instructionbox_stage}[ Concept Alignment Instruction]

You are a table-aware logic mapper with clinical reasoning ability.  

You are given:  

- A logic description in natural language, generated during solution planning, describing what needs to be computed and how.  

- A structured clinical table in TSV format, containing actual column names and sample values from the patient’s record.  

Your task is to align the logic with the table structure by rewriting the logic description using only the exact column names and observed values from the table.  

\textbf{General Guidelines:}  

\begin{itemize}
  \item Carefully read the table’s column names and value examples.  

  \item Replace all approximate, generic, or synonymous field references (e.g., “patient ID”, “disorder name”, “temperature”) with exact column names from the table.  

  \item Replace any approximate value references (e.g., “female”, “white”, “elevated”) with exact values from the table if they are semantically matched.  

  \item If no match can be found for a field or value, leave it unchanged — do not guess or hallucinate.  

  \item Preserve structural keywords and logical phrasing (e.g., “return”, “filter”, “if… then…”).  
\end{itemize}  

\textbf{Clinical-Specific Guidance:}  

\begin{itemize}
  \item If the logic involves detecting clinical conditions, map condition names to exact codes (e.g., SNOMED-CT or ICD) found in the table.  

  \item If the logic involves symptoms, vital signs, or lab results, identify the relevant fields (e.g., DESCRIPTION, VALUE, UNITS) where those observations may appear.  

  \item If the logic involves treatment recommendation, also identify signs that may indicate contraindications (e.g., allergy, rash, hypersensitivity).  
\end{itemize}  

\textbf{Important Constraints:}  

\begin{itemize}
  \item Do NOT invent or assume any column names or values that are not present in the table.  

  \item Do NOT simplify or paraphrase the logic — preserve its full structure and clinical meaning.  

  \item Do NOT output code, JSON, or any final prediction. Only return the rewritten logic as full English sentences.  
\end{itemize}  

Your output should be a rewritten version of the original logic, fully aligned to the specific column names and values in the table, while preserving the intended reasoning process.  

\end{instructionbox_stage}

\subsection{Adaptive Execution}

In this final stage, EHRMaster determines how to execute the aligned logic to obtain the answer.  

Depending on the nature of the task and the logic formulation, the system decides whether to generate executable code (e.g., Python or SQL) or to perform reasoning directly over the aligned logic.  For example, arithmetic computation tasks may be best handled via code execution, while simpler boolean checks may be performed through direct language-based reasoning.  

This stage consists of two steps:  
First, the system predicts whether code execution is needed.  
Second, based on this decision, the model is prompted with the appropriate instruction to produce the final answer.  

We design two distinct execution instructions accordingly.  If code execution is selected, the model generates Python code based on the aligned logic, which is then executed by a Python interpreter to compute the final answer from the structured input.  If direct reasoning is selected, the model performs the logic in natural language and outputs the final answer directly.

Both instruction formats are shown below.

\begin{instructionbox_stage}[Code Execution Instruction]

You are a Python coding assistant.  

You are given:  

- A natural language instruction describing the task objective.  

- A structured logic description aligned with table fields (produced from the previous alignment step).  

- A pandas DataFrame named \texttt{df}, already loaded in memory.  

- The DataFrame contains actual column names and sample values.  

Your goal is to write Python code that executes the described logic over the DataFrame to obtain the final result.  

\vspace{1mm}
\textbf{General Requirements:}

\begin{itemize}
  \item Use only the variable \texttt{df}.  
  \item Do NOT include import statements (e.g., no \texttt{import pandas}).  
  \item Do NOT redefine or reload the DataFrame.  
  \item Only use column names that exist in the DataFrame.  
  \item Assign the final result to a variable named \texttt{result}.  
  \item Do NOT include any explanation, markdown, or formatting (e.g., no \texttt{\textbackslash\textbackslash python}).  
  \item Write all code at the top level — do not wrap code in functions or conditional blocks.  
\end{itemize}

\vspace{1mm}
\textbf{Task-Specific Guidelines:}

\begin{itemize}
  \item For simple retrieval tasks, one line of filtering may be sufficient.  

  \item For reasoning tasks involving filtering, aggregation, or computation, you may use multiple lines of code.  
  Use intermediate variables as needed (e.g., \texttt{filtered\_df}, \texttt{avg\_value}).  

  \item If the task involves clinical codes (e.g., diagnosis or procedure codes), ensure proper type conversion before string operations.  
  For example, if the column \texttt{CODE} contains integers, use: \texttt{df['CODE'] = df['CODE'].astype(str)}.  
\end{itemize}

You must faithfully implement the logic as described, using only the provided table structure and column names.  
Only output executable Python code — no commentary, no natural language.  

\end{instructionbox_stage}

\begin{instructionbox_stage}[Direct Reasoning Instruction]

You are a clinical reasoning assistant.  

You are given:  

- A natural language instruction that defines the binary decision task.  

- A logic statement aligned with the table's column names and observed values.  

- A structured table (in TSV format) containing patient-specific clinical records.  

Your goal is to examine the table and reasoning logic to determine whether the target condition holds true.  

\vspace{1mm}
\textbf{Guidelines:}

\begin{itemize}
  \item Apply the logic to the table rows to assess whether the described condition is met.  

  \item Consider all relevant fields, including symptom descriptions, measurement values, diagnosis indicators, or other observed patterns.  

  \item Your output must be a binary decision (0 or 1), indicating whether the logic conditions are satisfied.  
\end{itemize}

\vspace{1mm}
\textbf{Constraints:}

\begin{itemize}
  \item Return your answer in the format: \texttt{Label: 0} or \texttt{Label: 1}.  
  \item Do NOT include any explanation, justification, or additional text.  

  \item VALUE fields may contain numeric or textual data — handle both robustly.  

  \item Be tolerant to incomplete or noisy entries.  
\end{itemize}

You must only return the prediction line, strictly following the expected format.  

\end{instructionbox_stage}

\section{Detailed Results}\label{sec:detailed_results}

\subsection{Full Results of Input Format Comparison} \label{sec:format_comparison_appendix}

Table~\ref{tab:LLM-format-full} presents the comprehensive evaluation results of general large language models (LLMs) across 4 distinct input formats on all benchmark tasks. These formats include plain text conversion, special character separation, graph-structured representation, and natural language description.
While Section~\ref{sec:inputprompt} summarizes the key takeaways, these results provide a granular foundation for that analysis. The full table enables readers to observe format-specific performance variations across different LLMs and tasks. The most salient trends include:
\begin{itemize}
\item Natural language description often yields a notable performance improvement for Reasoning tasks under the Data-Driven scenario, a trend particularly pronounced in powerful models such as the Gemini and GPT series.
\item Graph-structured representations are particularly effective for Understanding tasks, likely by helping models better align with the inherent schema of the structured data.

\item Conversely, no single formatting strategy consistently enhances performance on Knowledge-Driven tasks. This highlights that success in this domain likely requires more advanced clinical alignment techniques beyond input representation.
\end{itemize}

Overall, these findings reinforce the importance of input representation in structured medical reasoning, especially when working with general LLMs.

\subsection{Extended Few-shot Results}
\label{app:fewshot}

\begin{figure}[htbp]
    \centering

    \begin{subfigure}{0.19\textwidth}
        \includegraphics[width=\linewidth]{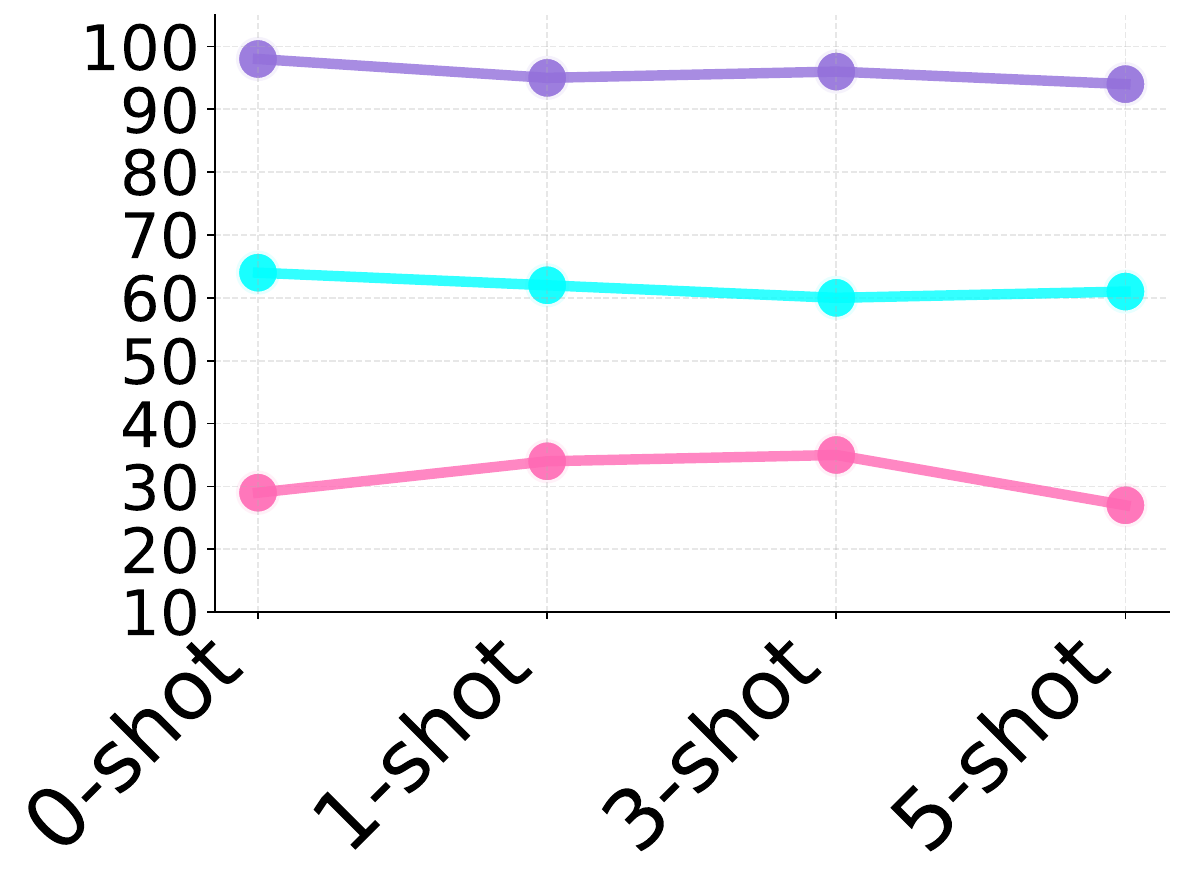}
        \subcaption{D-U1}
    \end{subfigure}
    \begin{subfigure}{0.19\textwidth}
        \includegraphics[width=\linewidth]{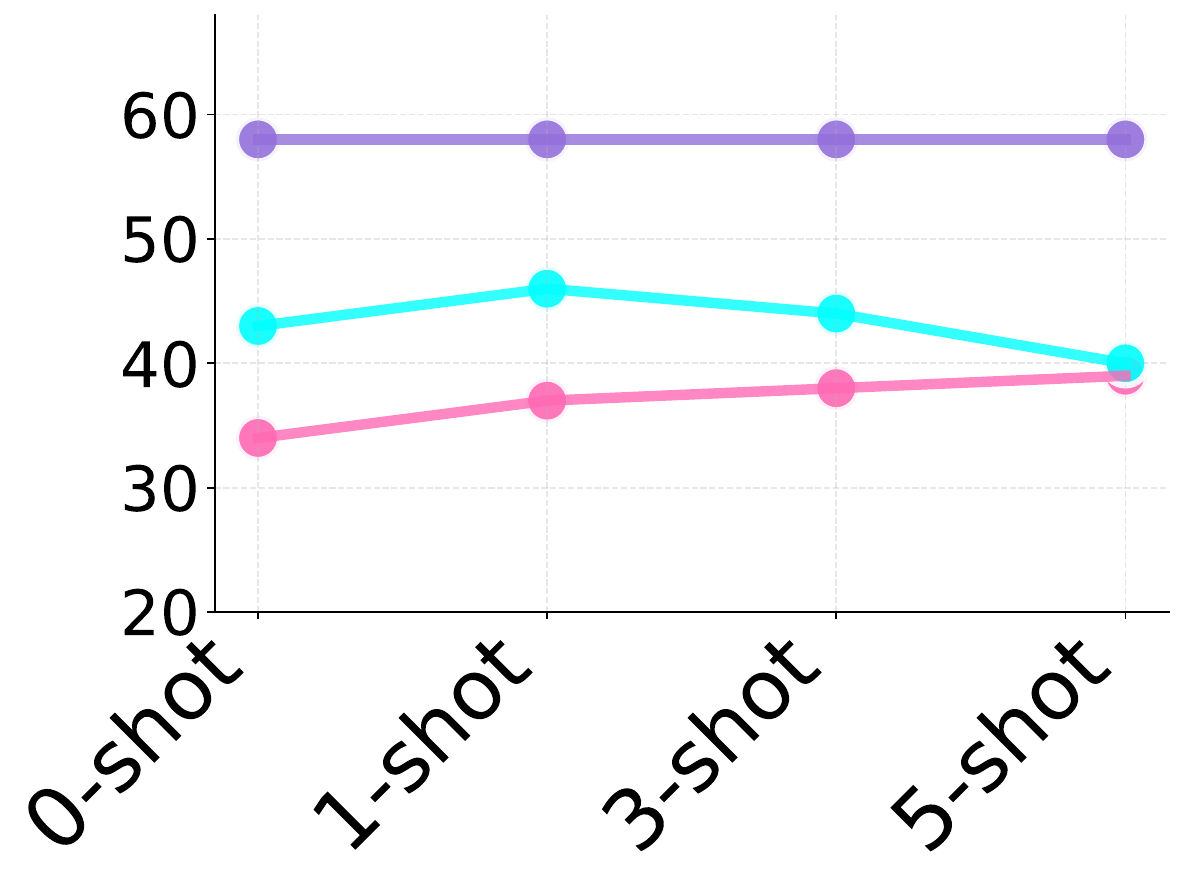}
        \subcaption{D-U2}
    \end{subfigure}
    \begin{subfigure}{0.19\textwidth}
        \includegraphics[width=\linewidth]{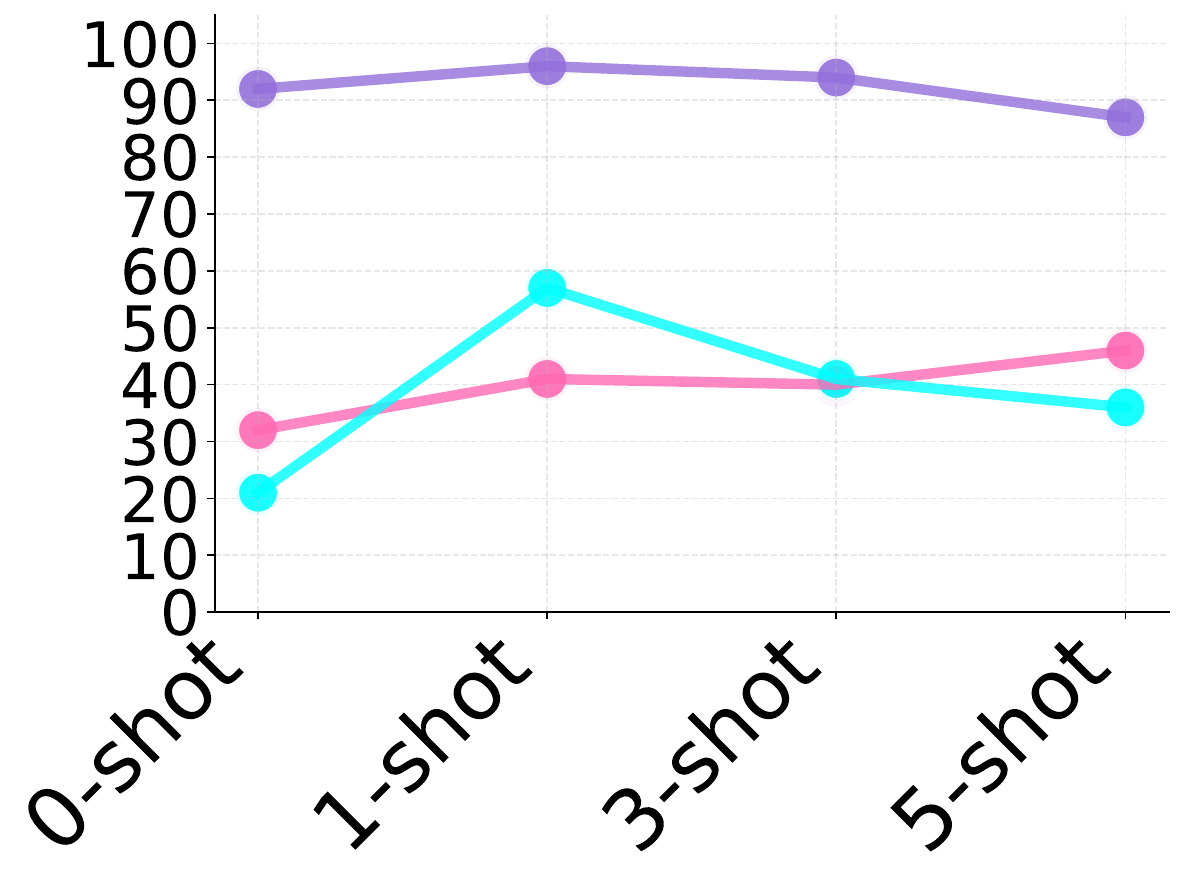}
        \subcaption{D-R1}
    \end{subfigure}
    \begin{subfigure}{0.19\textwidth}
        \includegraphics[width=\linewidth]{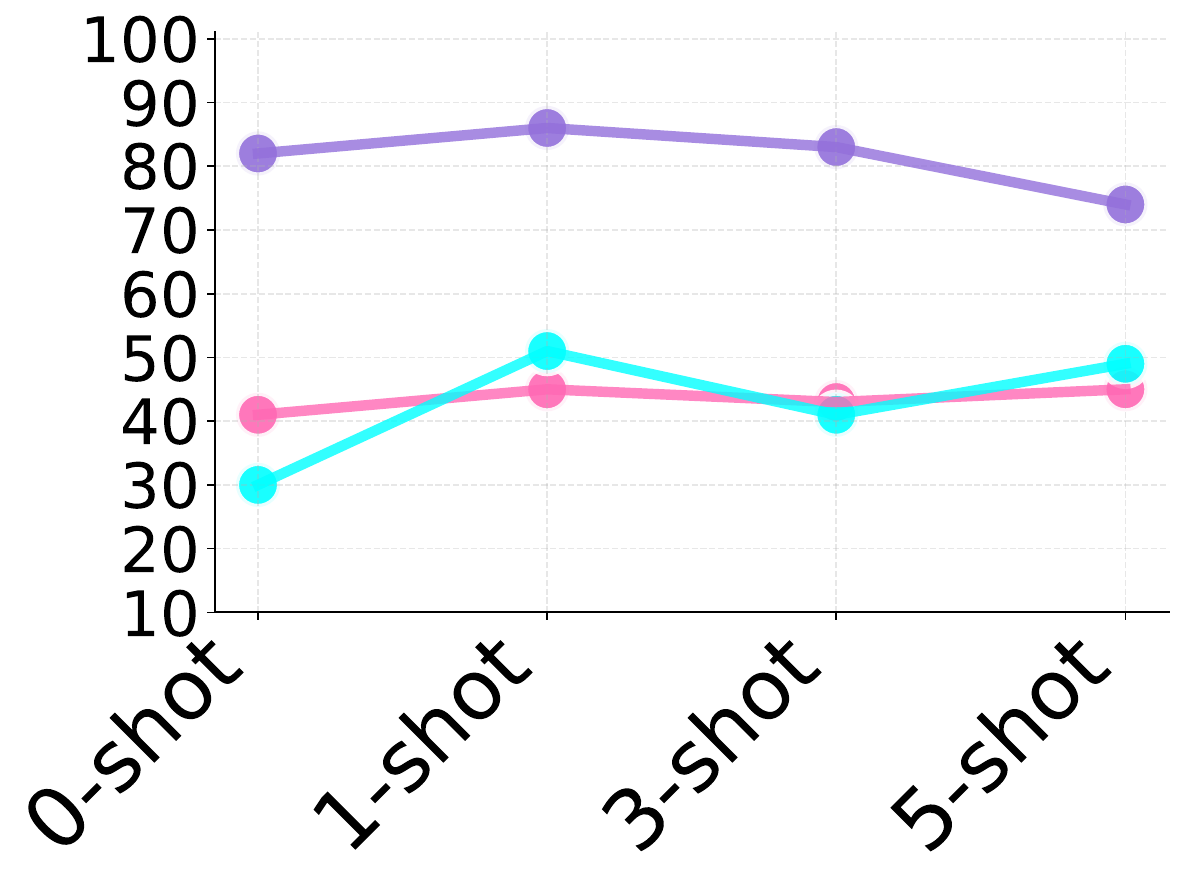}
        \subcaption{D-R2}
    \end{subfigure}
    \begin{subfigure}{0.19\textwidth}
        \includegraphics[width=\linewidth]{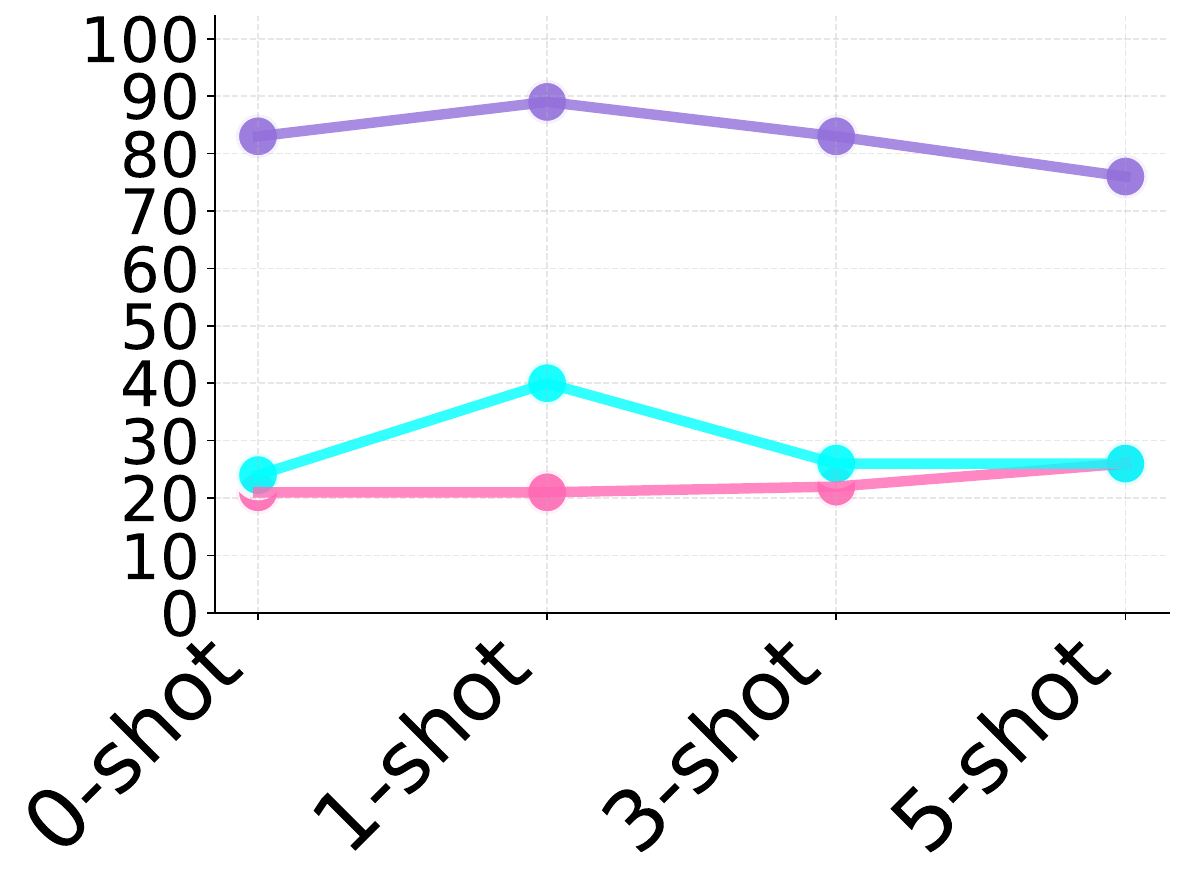}
        \subcaption{D-R3}
    \end{subfigure}
    \begin{subfigure}{0.19\textwidth}
        \includegraphics[width=\linewidth]{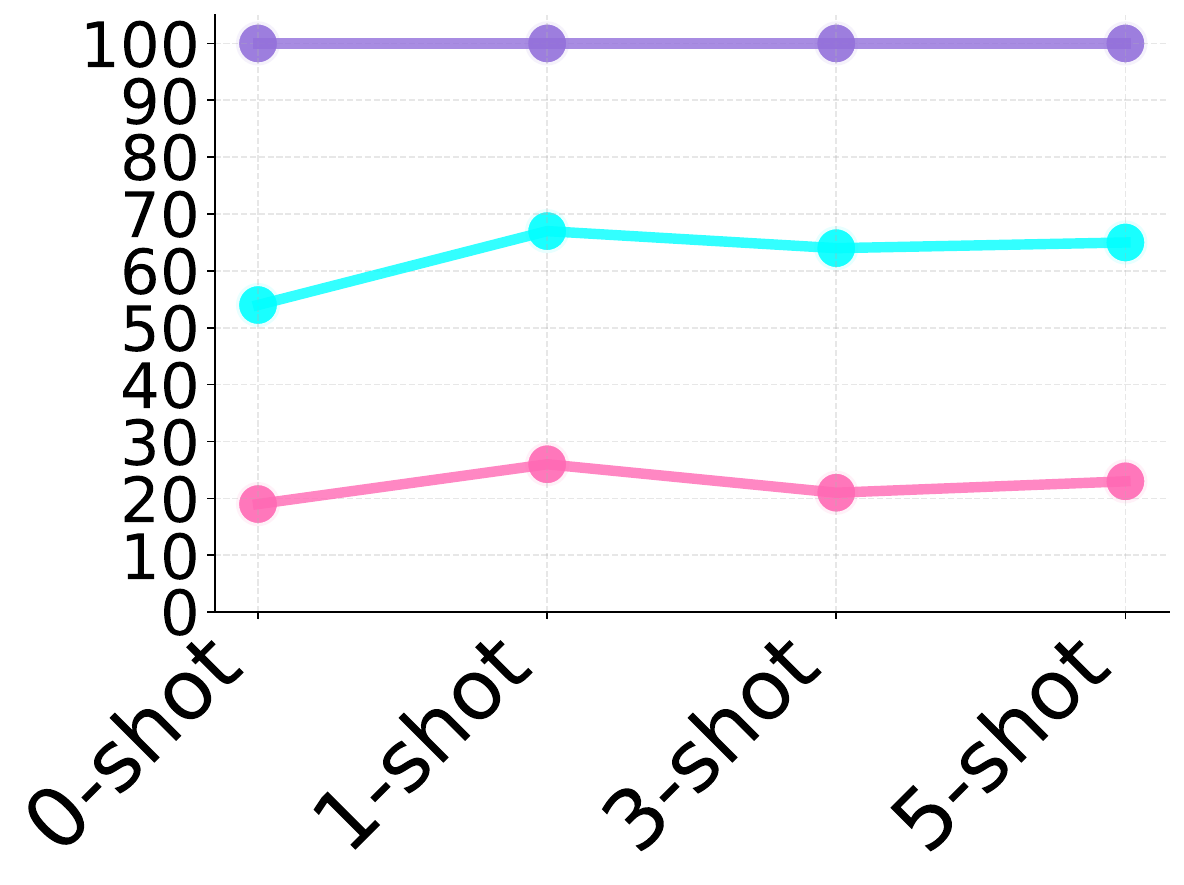}
        \subcaption{D-R4}
    \end{subfigure}
    \begin{subfigure}{0.19\textwidth}
        \includegraphics[width=\linewidth]{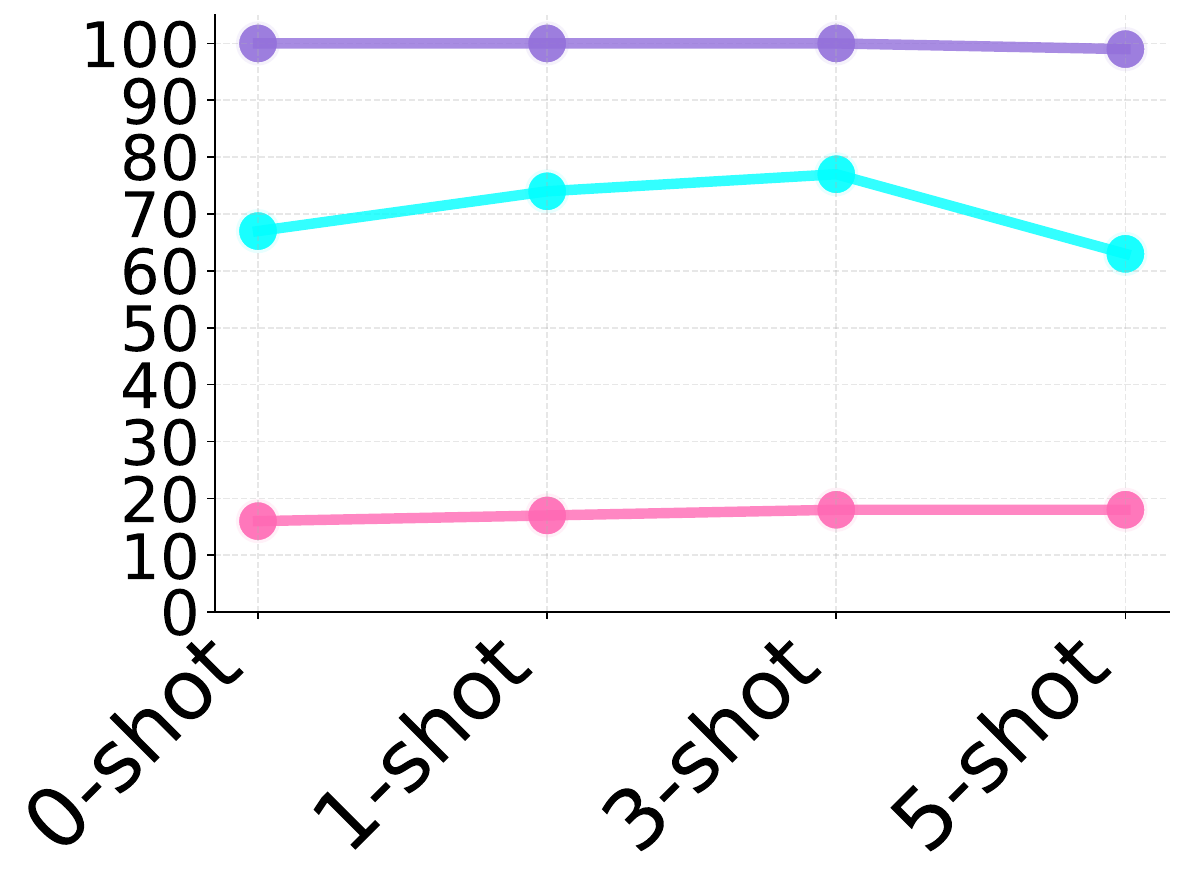}
        \subcaption{D-R5}
    \end{subfigure}

    \vspace{2mm} 

    \begin{subfigure}{0.19\textwidth}
        \includegraphics[width=\linewidth]{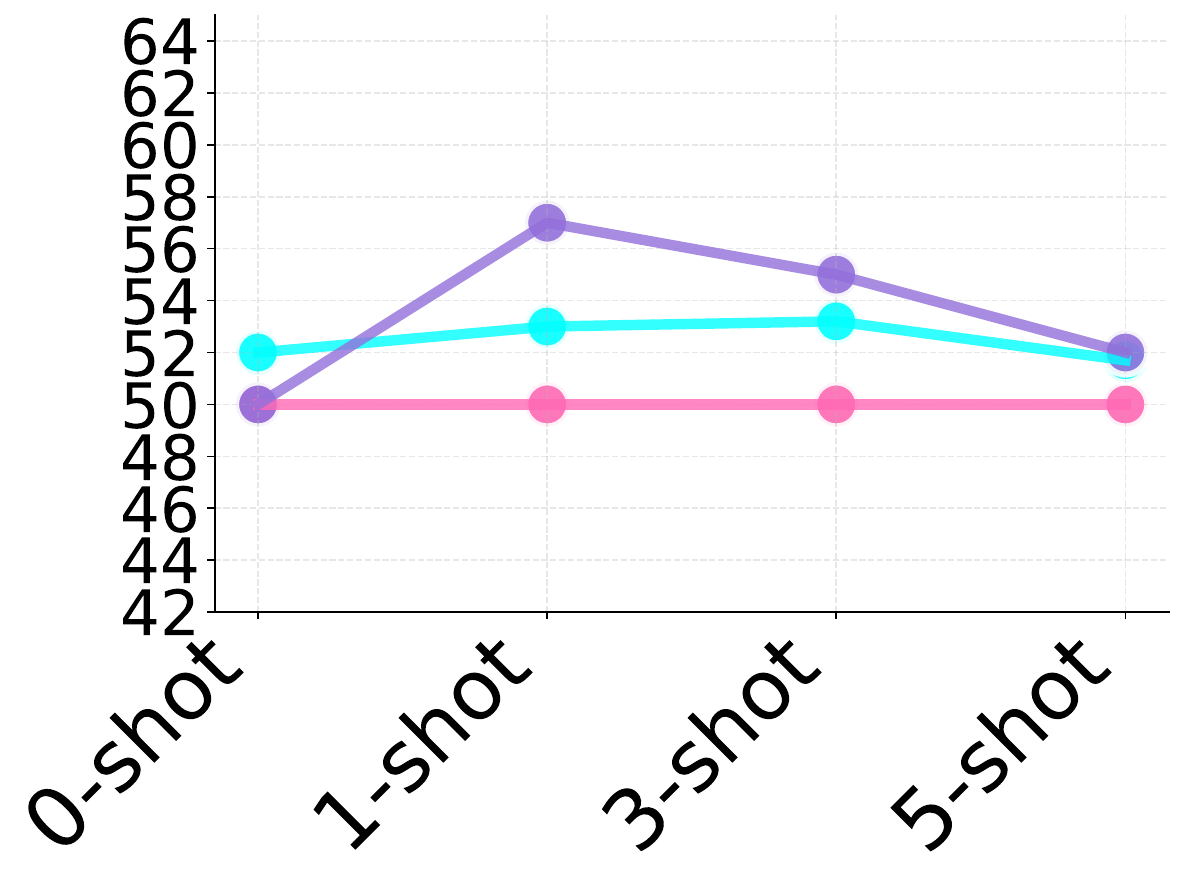}
        \subcaption{K-U1}
    \end{subfigure}
    \begin{subfigure}{0.19\textwidth}
        \includegraphics[width=\linewidth]{K-R1_few_shot.pdf}
        \subcaption{K-R1}
    \end{subfigure}
    \begin{subfigure}{0.19\textwidth}
        \includegraphics[width=\linewidth]{K-R2_few_shot.pdf}
        \subcaption{K-R2}
    \end{subfigure}
    \begin{subfigure}{0.19\textwidth}
        \includegraphics[width=\linewidth]{K-R3_few_shot.pdf}
        \subcaption{K-R3}
    \end{subfigure}
\raisebox{0.5ex}{\textcolor[HTML]{FF69B4}{\rule{0.8cm}{1pt}}} Gemini 1.5 \quad
\raisebox{0.5ex}{\textcolor[HTML]{00FFFF}{\rule{0.8cm}{1pt}}} Gemini-2.0 \quad
\raisebox{0.5ex}{\textcolor[HTML]{9370DB}{\rule{0.8cm}{1pt}}} Gemini-2.5
    \caption{Performance of LLMs on two Scenarios under few-shot (1, 3, and 5-shot) learning settings.}
    \label{fig:fewshot_detailed}
\end{figure}

Figure~\ref{fig:fewshot_detailed} presents the complete few-shot performance curves across all eleven tasks.

Several additional insights can be drawn:

\begin{itemize}
    \item \textbf{Task-specific variability:} Some tasks (e.g., D-R3, D-R5) exhibit a clear peak at 3-shot, while others (e.g., K-R3, K-R2) show flat or even degraded performance beyond 1-shot. This suggests task-specific sensitivity to the number of demonstrations.

    \item \textbf{Overfitting at 5-shot:} Across multiple tasks (e.g., D-R2, K-R3), the 5-shot setting yields slightly worse performance than 3-shot. This may reflect overfitting to a limited context window or interference between too many examples.

    \item \textbf{Modality difference:} Data-Driven tasks (D-*) often benefit more clearly from few-shot prompting than Knowledge-Driven tasks (K-*), indicating that in-context demonstrations help more with numerical pattern recognition than factual recall or symbolic reasoning.

    \item \textbf{Model-wise consistency:} Gemini-2.5 consistently outperforms Gemini-1.5 and 2.0 across nearly all tasks and shot settings, with particularly large margins in D-R1, D-R3, and K-R1. This confirms the robustness and improved sample efficiency of newer Gemini variants.

    \item \textbf{Marginal utility:} The performance gain from 0-shot to 1-shot is generally larger than that from 1-shot to 3-shot, implying diminishing returns with increasing few-shot size. This is especially noticeable in tasks like D-U1 and K-R2.
\end{itemize}

These extended observations highlight both the effectiveness and limitations of few-shot prompting in structured EHR tasks, reinforcing the need to calibrate shot size per task and model capability.

\subsection{Detailed Performance of LM-Based Enhancement
Methods on Structured EHR Tasks}
\label{sec:sota_full}

To supplement the analysis in Section~\ref{sec:sota}, we present the complete evaluation results of all 11 state-of-the-art methods across the 11 structured EHR tasks as follows: 

\begin{itemize}
    \item C.L.E.A.R.~\cite{deng2024enhancing} is a prompting-based baseline designed for temporal question answering over semi-structured tables, such as Wikipedia infoboxes. It enhances LLMs’ structured reasoning by guiding them through five steps—Comprehend, Locate, Examine, Analyze, Resolve—to extract and combine relevant tabular evidence, improving accuracy on time-related queries without requiring model fine-tuning.
    \item TaT~\cite{sun2025table} is a structured reasoning framework developed for planning and mathematical problem-solving tasks. Instead of generating free-form reasoning steps, it organizes LLM thoughts into a dynamic table where each row represents a step in the reasoning process and each column encodes constraints or sub-goals. By iteratively constructing and verifying this structured table, TaT enables large language models to better manage multi-step constraints and enhance logical consistency in structured reasoning tasks.
    \item TableMaster~\cite{cao2025tablemaster} is a training-based method developed for natural language question answering over web tables. It formulates table understanding as a sequence-to-sequence generation task and introduces structured recipes that supervise the model to perform layout-aware grounding, evidence selection, and answer decoding. By explicitly aligning question intent with table structure during fine-tuning, TableMaster significantly improves LLM performance on table-based QA benchmarks.
    \item TIDE~\cite{yang2025triples} is a decomposition-based method for complex question answering over web tables, focusing on improving multi-step reasoning. It restructures the input table into a set of symbolic triples, enabling LLMs to decompose questions into sub-questions and perform step-by-step verification using structured evidence. This explicit representation of table semantics helps LLMs better trace intermediate logic and enhances answer correctness in complex table QA.
    \item  E\textsuperscript{5}~\cite{zhang2024e5} is a zero-shot framework developed for hierarchical analysis over semi-structured tables, such as those found in financial or organizational reports. It augments LLMs with five structured operations—Explain, Extract, Execute, Exhibit, and Extrapolate—to guide complex reasoning without fine-tuning. Each operation handles a distinct subtask, from identifying key concepts to deriving trends, enabling LLMs to perform multi-layered analysis on real-world tabular data.
    \item  GraphOTTER~\cite{li2024graphotter} is a retrieval-augmented method developed for complex question answering over relational tables. It converts multi-table data into a dynamic entity-centric graph and performs multi-hop reasoning via a two-stage LLM pipeline: first retrieving evidence paths, then synthesizing final answers. This structure-aware design enables LLMs to handle compositional queries and cross-table dependencies more effectively than standard sequence-based methods.
    \item H-STAR~\cite{abhyankar2024h} is a hybrid reasoning framework developed for complex question answering over relational databases. It integrates SQL-based symbolic reasoning with text-based LLM inference through a two-stage process: selecting appropriate reasoning mode via LLM, then executing either a structured SQL query or a textual reasoning plan. By combining symbolic precision with LLM flexibility, H-STAR improves answer accuracy on table QA tasks involving aggregation, filtering, and multi-step logic.
    \item Table-R1~\cite{yang2025table} is an inference-time scaling method developed for numerical reasoning over structured tables, particularly in tasks requiring arithmetic or comparison. Without additional training, it boosts LLM performance by dynamically generating and verifying multiple reasoning paths through self-consistency and path pruning. This approach enables more accurate handling of complex table questions involving multi-step numerical logic.
    \item LLM4Healthcare~\cite{zhu2024prompting} is a prompting-based method designed for clinical prediction using structured longitudinal EHR data. It reformats patient records into textual prompts that describe temporal sequences of diagnoses, procedures, and medications, allowing general-purpose LLMs to perform zero-shot prediction tasks such as mortality or readmission. This approach demonstrates the feasibility of applying LLMs directly to structured medical data without fine-tuning.
    \item DeLLiriuM~\cite{contreras2024dellirium} is a fine-tuned LLM developed for delirium prediction in intensive care settings using structured EHR data. It converts time-series ICU records into flattened textual representations and fine-tunes a generative model on labeled prediction tasks. By aligning medical codes, timestamps, and outcomes into natural language, DeLLiriuM enhances clinical risk prediction within a language modeling framework.
    \item EnsembleLLM~\cite{hu2024power} is a multi-source prompting strategy designed for clinical decision support in oncology using structured EHR and model-derived features. It combines patient-level structured data with outputs from traditional machine learning models, presenting both as context to GPT-4o for downstream prediction tasks such as lymph node metastasis. This ensemble design leverages the complementary strengths of statistical models and LLMs to enhance interpretability and predictive accuracy in cancer care.
\end{itemize}

\subsection{Full Benchmark Results on Synthea and eICU} \label{sec:benchmark_appendix}

Table~\ref{tab:synthea_benchmark} and Table~\ref{tab:eicu_benchmark} present the complete evaluation results of 20 large language models (LLMs) on structured EHR tasks under the zero-shot setting, using two datasets: the synthetic Synthea dataset and the real-world eICU dataset. Both datasets span 11 tasks covering two clinical scenarios—Data-Driven and Knowledge-Driven—and two cognitive levels—Understanding (U) and Reasoning (R). Together, they provide a holistic view of model behavior under both controlled and realistic clinical conditions.

\paragraph{General-purpose LLMs consistently outperform medical-domain models.}

Across both the synthetic Synthea dataset (Table~\ref{tab:synthea_benchmark}) and the real-world eICU dataset (Table~\ref{tab:eicu_benchmark}), general-purpose LLMs—particularly Gemini-2.5 and GPT-4.1—consistently achieve higher performance than medical-domain models in nearly all tasks.

On the Synthea dataset, Gemini-2.5 achieves the highest accuracy in all seven Data-Driven tasks. It scores 98 on D-U1, 58 on D-U2, 92 on D-R1, 82 on D-R2, 83 on D-R3, and achieves perfect accuracy  on D-R4 and D-R5. It is also among the top-3 models in Knowledge-Driven tasks, achieving 58.7 AUC on K-R1 (1st) and 54.1 on K-R2 (3rd). However, it fails to produce valid outputs on K-U1 and K-R3. GPT-4.1 ranks 2nd on D-U1 (79), D-U2 (51), D-R1 (52), D-R2 (56) and D-R3 (48). It also produces the best AUC on K-U1 (55) and valid results across all Knowledge-Driven tasks, with AUCs of 55.6 (K-R1), 53.2 (K-R2), and 51 (K-R3).

On the eICU dataset, Gemini-2.5 ranks 1st on five of the seven Data-Driven tasks: D-U1 (95), D-U2 (84), D-R1 (95), D-R3 (36), and D-R5 (53). It ranks 4th on D-R2 (47, behind GPT-4.1, Gemini-1.5 and Gemini-2.0). It also achieves the best AUCs across all three Knowledge-Driven reasoning tasks: 53.2 on K-U1, 60.3 on K-R2, and 62.1 on K-R3, and places 2nd on K-R1 with 61.1. GPT-4.1 ranks 2nd on four Data-Driven tasks in eICU: D-U2 (52), D-R1 (52), D-R2 (67), and D-R4 (96). It ranks 1st on D-R3 (36), but drops to 11 on D-R5. It achieves the best AUC on K-R1 (63.3), slightly outperforming Gemini-2.5, and also produces valid outputs on K-U1 (50.7), K-R2 (56), and K-R3 (61.4).

In contrast, most medical-domain LLMs fail to produce valid outputs in the Knowledge-Driven setting. On the Synthea dataset, models such as Huatuo, HEAL, Meditron-7B, and PMC\_LLaMA\_13B yield no valid output on all four Knowledge-Driven tasks. Their performance in Data-Driven tasks is also low: for instance, MedAlpaca-13B scores no higher than 11 on any task, and Med42-70B reaches only 27 on D-R4. Similar patterns hold on the eICU dataset, where most medical models again fail to output any valid AUC and achieve accuracy below 25\% across nearly all Data-Driven tasks. None of the medical-domain models ranks in the top-3 for any task in either dataset.

\paragraph{Task difficulty is shaped by both reasoning complexity and dataset characteristics.}

Across both the Synthea and eICU datasets, task difficulty is primarily driven by reasoning complexity: Reasoning tasks are consistently more challenging than Understanding tasks. For example, in the Data-Driven setting, top-performing models like Gemini-2.5 and GPT-4.1 achieve high accuracy on Understanding tasks such as D-U1 and D-U2—reaching 98 and 58 (Synthea), and 95 and 84 (eICU) for Gemini-2.5. In contrast, their performance drops on Reasoning tasks. On D-R3, Gemini-2.5 scores 83 in Synthea but only 36 in eICU.

While this reasoning-related gap is consistent across datasets, the absolute performance also heavily depends on the dataset. For instance, DeepSeek-V3 scores 59 on D-R5 in eICU, compared to 90 in Synthea, showing large variance even under the same task design. These results suggest that real-world data introduces additional modeling challenges beyond reasoning complexity alone.

In the Knowledge-Driven setting, all tasks—including both Understanding (K-U1) and Reasoning (K-R1 to K-R3)—are substantially more difficult. Most models fail to produce any valid outputs on these tasks, especially those in the medical domain. Among general-purpose models, only a few achieve consistent AUC above 55. For example, Gemini-2.5 reaches 58.7 on K-R1 (Synthea) and 62.1 on K-R3 (eICU); GPT-4.1 achieves 61.4 on K-R3 (eICU) and 55.6 on K-R1 (Synthea). However, their performance still fluctuates with the dataset—for instance, GPT-4.1 scores 55 on K-U1 in Synthea but only 50.7 in eICU—again reflecting the compound impact of task type and dataset characteristics.

\paragraph{Medical-domain LLMs remain largely non-functional in Knowledge-Driven settings.}

Across both the Synthea and eICU datasets, nearly all medical-domain LLMs fail to produce valid outputs on Knowledge-Driven tasks. Specifically, models such as Huatuo, HEAL, Meditron-7B, PMC\_LLaMA\_13B, and MedAlpaca-13B output no valid AUC on any of the four Knowledge-Driven tasks in either dataset. Even the relatively stronger medical models, such as ClinicalCamel70B and Med42-70B, produce only one or two valid outputs across all Knowledge-Driven tasks, and their AUC scores remain below 50. In contrast, general-purpose LLMs like GPT-4.1 and Gemini-2.5 not only produce valid predictions on all Knowledge-Driven tasks but also achieve AUCs consistently above 50, with top scores reaching 62.1 on K-R3 in eICU.

This performance gap is likely rooted in the pretraining data and objective differences. Most medical-domain LLMs are trained primarily on unstructured clinical text, such as biomedical literature, discharge summaries, or doctor-patient conversations. These sources emphasize natural language fluency and domain-specific terminology but provide limited exposure to structured representations like tables, field-value pairs, or code-based logic. As a result, such models lack the inductive bias needed to align and reason over structured EHR inputs. Consequently, they struggle with tasks that require multi-field alignment, code mapping, or temporal reasoning—even in domains where their medical knowledge should be advantageous.

\subsection{Finetuning Analysis} 

Figure~\ref{fig:finetune_all} shows that finetuning substantially improves model performance across both Data-Driven and Knowledge-Driven tasks. Across all tasks, multi-task finetuning consistently outperforms single-task finetuning, likely because joint training helps the model learn shared structures and reasoning patterns relevant to structured EHR data.

In the Data-Driven setting (Figure~\ref{fig:finetune_all}a), performance gains are especially pronounced on reasoning tasks. For example, multi-task finetuning improves D-R2 and D-R3 by over 20 percentage points compared to the baseline, outperforming single-task finetuning by a wide margin. Even in simpler understanding tasks like D-U1 and D-U2, single-task finetuning yields modest gains, while multi-task variants further amplify these improvements. This suggests that structural patterns shared across Data-Driven tasks—such as field-level filtering, logical comparison, or numeric aggregation—can be effectively captured through joint optimization.

In the Knowledge-Driven setting (Figure~\ref{fig:finetune_all}b), the relative improvements are smaller in absolute terms but still meaningful. On K-R1 and K-R2, multi-task finetuning improves AUC by over 2 points compared to the baseline and outperforms single-task training by 1–2 points. However, on K-R3, all methods struggle to achieve substantial gains, indicating that this task may involve more complex or dataset-specific reasoning that is not easily generalized from other tasks.

Overall, these results demonstrate that even with limited data, parameter-efficient finetuning can enhance performance on structured EHR tasks—particularly when training is conducted jointly across multiple task types. The consistent advantage of multi-task training suggests that cross-task inductive transfer plays an important role in helping models generalize reasoning over structured inputs, especially when supervision is scarce.

\definecolor{pastelBlue}{HTML}{A6D0E4}

\begin{table*}[htbp]
\begin{tcolorbox}[colback=pastelBlue!10, colframe=pastelBlue!80!black, arc=3mm, boxrule=1pt]
\centering
\setlength{\tabcolsep}{4pt}
\renewcommand{\arraystretch}{1.25}

\resizebox{\textwidth}{!}{%
\begin{tabular}{ll
                cc        
                ccccc     
                c         
                ccc       
                }
\toprule
\multirow{3}{*}{\bfseries Model} & \multirow{3}{*}{\bfseries Format} 
& \multicolumn{7}{c}{\bfseries Data-Driven} 
& \multicolumn{4}{c}{\bfseries Knowledge-Driven} \\
\cmidrule(lr){3-9} \cmidrule(lr){10-13}
& & \multicolumn{2}{c}{Understanding} & \multicolumn{5}{c}{Reasoning} 
  & \multicolumn{1}{c}{Understanding} & \multicolumn{3}{c}{Reasoning} \\
& & D-U1 & D-U2 & D-R1 & D-R2 & D-R3 & D-R4 & D-R5 
  & K-U1 & K-R1 & K-R2 & K-R3 \\
\midrule
GPT-3.5 Turbo & plain text conversion 
& 6 & 15 & 14 & 18 & 7 & 7 & 24 
& \textcolor{red}{\ding{55}} & 58.1 & 55.4 & 52.9 \\
GPT-3.5 Turbo & special character separation 
& \textbf{8} & 14 & \textbf{20} & 25 & \textbf{18} & 3 & 7 
& \textbf{51} & \textbf{59.3} & 52.2 & \textbf{60.2} \\
GPT-3.5 Turbo & graph-structured representation 
& 7 & \textbf{28} & 10 & 19 & 12 & 3 & 8 
& \textcolor{red}{\ding{55}} & 53.1 & 53.1 & \textcolor{red}{\ding{55}} \\
GPT-3.5 Turbo & natural language description  
& 5 & 16 & 15 & \textbf{26} & 14 & \textbf{11} & \textbf{59} 
& \textcolor{red}{\ding{55}} & 50.9 & \textbf{57.7} & \textcolor{red}{\ding{55}} \\
\midrule
GPT-4.1 & plain text conversion 
& \textbf{79} & 51 & \textbf{52} & 56 & \textbf{48} & \textbf{70} & \textbf{84} 
& \textbf{55} & \textbf{55.6} & 53.2 & 51 \\
GPT-4.1 & special character separation 
& 75 & 50 & 43 & 50 & 43 & 65 & 70 
& \textcolor{red}{\ding{55}} & 53.3 & 54.5 & \textbf{55.6} \\
GPT-4.1 & graph-structured representation 
& 58 & \textbf{54} & 48 & \textbf{60} & 47 & 66 & 80 
& \textcolor{red}{\ding{55}} & 54 & 50.7 & \textcolor{red}{\ding{55}} \\
GPT-4.1 & natural language description  
& 78 & 51 & 48 & 57 & 45 & 68 & 81 
& \textcolor{red}{\ding{55}} & 55 & \textbf{54.9} & \textcolor{red}{\ding{55}} \\
\midrule
Gemini 1.5 & plain text conversion 
& 29 & 34 & 32 & \textbf{41} & \textbf{21} & 19 & 16 
& \textcolor{red}{\ding{55}} & 55.6 & \textcolor{red}{\ding{55}} & \textcolor{red}{\ding{55}} \\
Gemini 1.5 & special character separation 
& 27 & 40 & 21 & 28 & 13 & 20 & 17 
& \textcolor{red}{\ding{55}} & 55.6 & \textcolor{red}{\ding{55}} & \textcolor{red}{\ding{55}} \\
Gemini 1.5 & graph-structured representation 
& \textbf{39} & \textbf{42} & \textbf{36} & 39 & 17 & 12 & 1 
& \textcolor{red}{\ding{55}} & \textbf{58.3} & \textcolor{red}{\ding{55}} & \textcolor{red}{\ding{55}} \\
Gemini 1.5 & natural language description  
& 28 & 33 & \textbf{36} & 31 & 10 & \textbf{38} & \textbf{37} 
& \textbf{51} & 52.8 & \textcolor{red}{\ding{55}} & \textcolor{red}{\ding{55}} \\
\midrule
Gemini 2.0 & plain text conversion 
& 64 & 43 & 21 & 30 & 24 & 54 & 67 
& 52 & 57.7 & \textbf{56.2} & 51.6 \\
Gemini 2.0 & special character separation 
& \textbf{77} & \textbf{50} & 20 & 32 & 26 & 62 & 70 
& \textbf{53} & 60.0 & 54.1 & \textcolor{red}{\ding{55}} \\
Gemini 2.0 & graph-structured representation 
& 72 & 46 & 8 & 16 & 10 & 20 & 30 
& \textcolor{red}{\ding{55}} & \textbf{61.1} & 53.1 & \textcolor{red}{\ding{55}} \\
Gemini 2.0 & natural language description  
& 57 & 40 & \textbf{25} & \textbf{37} & \textbf{35} & \textbf{84} & \textbf{82} 
& 51 & 56.2 & 54.1 & \textbf{52.2} \\
\midrule
Gemini 2.5 & plain text conversion 
& \textbf{98} & \textbf{58} & 92 & 82 & 83 & \textbf{100} & \textbf{100} 
& \textcolor{red}{\ding{55}} & \textbf{58.7} & 54.1 & \textcolor{red}{\ding{55}} \\
Gemini 2.5 & special character separation 
& 95 & 57 & 88 & 84 & 81 & 98 & \textbf{100} 
& \textcolor{red}{\ding{55}} & 57.5 & \textbf{55.1} & \textcolor{red}{\ding{55}} \\
Gemini 2.5 & graph-structured representation 
& \textbf{98} & 57 & 77 & 80 & 75 & \textbf{100} & 98 
& \textbf{52} & 56.8 & 54.1 & \textbf{51.6} \\
Gemini 2.5 & natural language description  
& 96 & \textbf{58} & \textbf{98} & \textbf{87} & \textbf{86} & 76 & 77 
& \textbf{52} & 58.1 & 53.1 & 51.0 \\
\midrule
DeepSeek-V2.5 & plain text conversion 
& \textbf{72} & \textbf{41} & \textbf{18} & \textbf{51} & 14 & 44 & 52 
& 51 & \textcolor{red}{\ding{55}} & \textcolor{red}{\ding{55}} & \textcolor{red}{\ding{55}} \\
DeepSeek-V2.5 & special character separation 
& \textbf{72} & 37 & 13 & 42 & 15 & 47 & 56 
& \textcolor{red}{\ding{55}} & \textbf{52.8} & \textcolor{red}{\ding{55}} & \textcolor{red}{\ding{55}} \\
DeepSeek-V2.5 & graph-structured representation 
& 26 & 19 & 14 & 41 & 13 & 28 & 42 
& \textbf{52} & \textcolor{red}{\ding{55}} & \textcolor{red}{\ding{55}} & \textcolor{red}{\ding{55}} \\
DeepSeek-V2.5 & natural language description  
& 65 & 39 & 13 & 49 & \textbf{16} & \textbf{74} & \textbf{73} 
& \textcolor{red}{\ding{55}} & \textcolor{red}{\ding{55}} & \textbf{54.1} & \textcolor{red}{\ding{55}} \\
\midrule
DeepSeek-V3 & plain text conversion 
& 72 & \textbf{41} & 8 & 37 & 12 & 72 & 90 
& \textcolor{red}{\ding{55}} & 52.8 & \textcolor{red}{\ding{55}} & 56.4 \\
DeepSeek-V3 & special character separation 
& 66 & 40 & 9 & 42 & 10 & 69 & 84 
& \textcolor{red}{\ding{55}} & \textbf{55.6} & \textcolor{red}{\ding{55}} & \textcolor{red}{\ding{55}} \\
DeepSeek-V3 & graph-structured representation 
& \textbf{72} & 37 & \textbf{13} & \textbf{47} & \textbf{14} & 76 & 89 
& \textcolor{red}{\ding{55}} & \textbf{55.6} & \textcolor{red}{\ding{55}} & \textcolor{red}{\ding{55}} \\
DeepSeek-V3 & natural language description  
& 65 & 39 & 11 & 38 & 12 & \textbf{88} & \textbf{96} 
& \textcolor{red}{\ding{55}} & 52.8 & \textbf{51.0} & \textcolor{red}{\ding{55}} \\
\midrule
Qwen-7B & plain text conversion 
& 1 & 7 & 4 & \textbf{24} & 1 & \textcolor{red}{\ding{55}} & \textcolor{red}{\ding{55}} 
& \textcolor{red}{\ding{55}} & \textcolor{red}{\ding{55}} & \textcolor{red}{\ding{55}} & \textcolor{red}{\ding{55}} \\
Qwen-7B & special character separation 
& 1 & 8 & 4 & 20 & 3 & \textcolor{red}{\ding{55}} & \textcolor{red}{\ding{55}} 
& \textcolor{red}{\ding{55}} & \textcolor{red}{\ding{55}} & \textcolor{red}{\ding{55}} & \textcolor{red}{\ding{55}} \\
Qwen-7B & graph-structured representation 
& 1 & \textbf{12} & 4 & 19 & 2 & \textcolor{red}{\ding{55}} & \textcolor{red}{\ding{55}} 
& \textcolor{red}{\ding{55}} & \textcolor{red}{\ding{55}} & \textcolor{red}{\ding{55}} & \textcolor{red}{\ding{55}} \\
Qwen-7B & natural language description  
& \textbf{2} & 9 & \textbf{8} & 16 & \textbf{4} & \textbf{12} & \textbf{13} 
& \textcolor{red}{\ding{55}} & \textcolor{red}{\ding{55}} & \textcolor{red}{\ding{55}} & \textcolor{red}{\ding{55}} \\
\midrule
Qwen-14B & plain text conversion 
& 4 & \textbf{30} & \textbf{19} & 17 & \textbf{11} & 16 & 4 
& \textcolor{red}{\ding{55}} & \textcolor{red}{\ding{55}} & \textcolor{red}{\ding{55}} & \textcolor{red}{\ding{55}} \\
Qwen-14B & special character separation 
& \textbf{5} & 29 & 17 & 14 & \textbf{11} & 16 & 6 
& \textbf{52} & \textcolor{red}{\ding{55}} & \textcolor{red}{\ding{55}} & \textcolor{red}{\ding{55}} \\
Qwen-14B & graph-structured representation 
& \textbf{5} & 28 & 14 & \textbf{21} & 9 & 13 & 2 
& \textcolor{red}{\ding{55}} & \textcolor{red}{\ding{55}} & \textcolor{red}{\ding{55}} & \textcolor{red}{\ding{55}} \\
Qwen-14B & natural language description  
& 2 & 22 & 18 & 13 & 8 & \textbf{44} & \textbf{35} 
& \textcolor{red}{\ding{55}} & \textcolor{red}{\ding{55}} & \textcolor{red}{\ding{55}} & \textcolor{red}{\ding{55}} \\
\midrule
Qwen-32B & plain text conversion 
& 25 & \textbf{25} & 24 & 26 & 15 & 47 & 10 
& \textcolor{red}{\ding{55}} & \textbf{58.3} & \textbf{51} & \textcolor{red}{\ding{55}} \\
Qwen-32B & special character separation 
& 26 & 23 & \textbf{26} & 32 & \textbf{16} & 51 & 18 
& \textcolor{red}{\ding{55}} & 55.6 & \textbf{51} & \textcolor{red}{\ding{55}} \\
Qwen-32B & graph-structured representation 
& \textbf{38} & \textbf{25} & 20 & 28 & 14 & 58 & 14 
& \textcolor{red}{\ding{55}} & \textbf{58.3} & \textbf{51} & \textcolor{red}{\ding{55}} \\
Qwen-32B & natural language description  
& 32 & 22 & 22 & \textbf{35} & \textbf{16} & \textbf{72} & \textbf{45} 
& \textcolor{red}{\ding{55}} & 55.6 & \textbf{51} & \textcolor{red}{\ding{55}} \\
\midrule
Qwen-72B & plain text conversion 
& 15 & 6 & 27 & \textbf{48} & 20 & 41 & 29 
& \textcolor{red}{\ding{55}} & \textcolor{red}{\ding{55}} & \textcolor{red}{\ding{55}} & 52.2 \\
Qwen-72B & special character separation 
& 18 & 6 & 23 & 42 & 16 & 46 & 34 
& \textcolor{red}{\ding{55}} & \textcolor{red}{\ding{55}} & \textcolor{red}{\ding{55}} & \textcolor{red}{\ding{55}} \\
Qwen-72B & graph-structured representation 
& 16 & 7 & \textbf{29} & 39 & \textbf{25} & 31 & 17 
& \textcolor{red}{\ding{55}} & \textcolor{red}{\ding{55}} & \textcolor{red}{\ding{55}} & \textcolor{red}{\ding{55}} \\
Qwen-72B & natural language description  
& \textbf{21} & \textbf{9} & 27 & 31 & 18 & \textbf{74} & \textbf{64} 
& \textcolor{red}{\ding{55}} & \textcolor{red}{\ding{55}} & \textcolor{red}{\ding{55}} & \textcolor{red}{\ding{55}} \\
\bottomrule
\end{tabular}}
\end{tcolorbox}
\caption{
Performance of general LLMs across 4 input formats on both Data-Driven and Knowledge-Driven tasks under the zero-shot setting. 
Bold values indicate the best performance for each task; \textcolor{red}{\ding{55}} indicates missing or inapplicable data.
}
\label{tab:LLM-format-full}
\end{table*}

\begin{table*}[htbp]
\centering
\begin{tabular}{ll*{7}{c}*{5}{c}}
  \toprule
  \multirow{4}{*}{\bfseries Types} & \multirow{4}{*}{\bfseries Models} 
  & \multicolumn{7}{c}{\bfseries Data-Driven } 
  & \multicolumn{5}{c}{\bfseries Knowledge-Driven} \\
  \cmidrule(lr){3-9} \cmidrule(lr){10-13}
  & & \multicolumn{2}{c}{\bfseries U (\,\%)} & \multicolumn{5}{c}{\bfseries R (\,\%)} 
  & \multicolumn{1}{c}{\bfseries U (\,\%)} & \multicolumn{3}{c}{\bfseries R (\,\%)} \\
  & & D-U1 & D-U2 & D-R1 & D-R2 & D-R3 & D-R4 & D-R5 
  & K-U1 & K-R1 & K-R2 & K-R3 \\
  \cmidrule(lr){3-14}
  & & ACC & ACC & ACC & ACC & ACC & ACC & ACC 
  & AUC & AUC  & AUC & AUC\\
  \midrule
  \multirow{10}{*}{\rotatebox{90}{\bfseries General LLMs}} & GPT-3.5 Turbo  & 36 & 5 & 6 & 36 & 2 & 21 & 3 & \textcolor{red}{\ding{55}}  & 52.9 & 55.8 & 55.8 \\
  & GPT-4.1 & 62 & \textbf{\textcolor[HTML]{F56B00}{52}} & \textbf{\textcolor[HTML]{F56B00}{52}} & \textbf{\textcolor[HTML]{F56B00}{67}} & \textbf{\textcolor[HTML]{3166FF}{36}} & \textbf{\textcolor[HTML]{F56B00}{96}} & 11 & \textbf{\textcolor[HTML]{32CB00}{50.7}}  & \textbf{\textcolor[HTML]{3166FF}{63.3}} & 56 & \textbf{\textcolor[HTML]{F56B00}{61.4}} \\
  \cmidrule(lr){2-13}
  & Gemini 1.5  & 53 & 27 & \textbf{\textcolor[HTML]{32CB00}{37}} & \textbf{\textcolor[HTML]{32CB00}{63}} & \textbf{\textcolor[HTML]{32CB00}{25}} & 74 & 15 & \textcolor{red}{\ding{55}}  & \textcolor{red}{\ding{55}} & \textbf{\textcolor[HTML]{32CB00}{56.1}} & \textcolor{red}{\ding{55}} \\
  & Gemini-2.0  & 34 & \textbf{\textcolor[HTML]{32CB00}{39}} & 35 & \textbf{\textcolor[HTML]{3166FF}{83}} & 24 & 59 & 45 & \textcolor{red}{\ding{55}}  & 57.6 & \textbf{\textcolor[HTML]{F56B00}{57.2}} & \textbf{\textcolor[HTML]{32CB00}{59.9}} \\
  & Gemini 2.5 & \textbf{\textcolor[HTML]{3166FF}{95}} & \textbf{\textcolor[HTML]{3166FF}{84}} & \textbf{\textcolor[HTML]{3166FF}{95}} & 47 & \textbf{\textcolor[HTML]{3166FF}{36}} & \textbf{\textcolor[HTML]{3166FF}{97}} & \textbf{\textcolor[HTML]{F56B00}{53}} & \textbf{\textcolor[HTML]{3166FF}{53.2}}  & \textbf{\textcolor[HTML]{32CB00}{61.1}} & \textbf{\textcolor[HTML]{3166FF}{60.3}} & \textbf{\textcolor[HTML]{3166FF}{62.1}} \\
    \cmidrule(lr){2-13}
  & DeepSeek-V2.5 & \textbf{\textcolor[HTML]{F56B00}{72}} & 41 & 18 & 47 & 14 & 44 & \textbf{\textcolor[HTML]{32CB00}{52}} & \textbf{\textcolor[HTML]{3166FF}{53.2}}  & \textcolor{red}{\ding{55}} & \textcolor{red}{\ding{55}} & \textcolor{red}{\ding{55}} \\
  & DeepSeek-V3 & \textbf{\textcolor[HTML]{F56B00}{72}} & 41 & 8 & 61 & 12 & \textbf{\textcolor[HTML]{32CB00}{72}} & \textbf{\textcolor[HTML]{3166FF}{59}} & \textcolor{red}{\ding{55}}  & 52.8 & \textcolor{red}{\ding{55}} & \textcolor{red}{\ding{55}} \\
    \cmidrule(lr){2-13}
  & Qwen-7B & 1 & 7 & 4 & 24 & 1 & \textcolor{red}{\ding{55}} & \textcolor{red}{\ding{55}} & \textcolor{red}{\ding{55}}  & \textcolor{red}{\ding{55}} & \textcolor{red}{\ding{55}} & \textcolor{red}{\ding{55}} \\
  & Qwen-14B & 4 & 30 & 19 & 17 & 11 & 16 & 4 & \textcolor{red}{\ding{55}}  & \textcolor{red}{\ding{55}} & \textcolor{red}{\ding{55}} & \textcolor{red}{\ding{55}} \\
  & Qwen-32B & 25 & 25 & 24 & 26 & 15 & 47 & 10 & \textcolor{red}{\ding{55}}  & \textbf{\textcolor[HTML]{F56B00}{58.3}} & 51 & \textcolor{red}{\ding{55}} \\
  & Qwen-72B & 15 & 6 & 27 & 48 & 20 & 41 & 29 & \textcolor{red}{\ding{55}}  & \textcolor{red}{\ding{55}} & \textcolor{red}{\ding{55}} & 52.2 \\
  \midrule
  \multirow{9}{*}{\rotatebox{90}{\bfseries Medical LLMs}} & Huatuo   & 2 & \textcolor{red}{\ding{55}} & \textcolor{red}{\ding{55}} & \textcolor{red}{\ding{55}} & \textcolor{red}{\ding{55}} & \textcolor{red}{\ding{55}} & \textcolor{red}{\ding{55}} & \textcolor{red}{\ding{55}}  & \textcolor{red}{\ding{55}} & \textcolor{red}{\ding{55}} & \textcolor{red}{\ding{55}} \\
  & HEAL & 10 & \textcolor{red}{\ding{55}} & \textcolor{red}{\ding{55}} & 7 & \textcolor{red}{\ding{55}} & \textcolor{red}{\ding{55}} & \textcolor{red}{\ding{55}} & \textcolor{red}{\ding{55}}  & \textcolor{red}{\ding{55}} & \textcolor{red}{\ding{55}} & \textcolor{red}{\ding{55}} \\
  & Meditron-7B & 3 & \textcolor{red}{\ding{55}} & \textcolor{red}{\ding{55}} & 11 & 3 & \textcolor{red}{\ding{55}} & \textcolor{red}{\ding{55}} & \textcolor{red}{\ding{55}}  & \textcolor{red}{\ding{55}} & \textcolor{red}{\ding{55}} & \textcolor{red}{\ding{55}} \\
    \cmidrule(lr){2-13}
  & MedAlpaca-13B & 11 & \textcolor{red}{\ding{55}} & \textcolor{red}{\ding{55}} & 10 & \textcolor{red}{\ding{55}} & 1 & \textcolor{red}{\ding{55}} & \textcolor{red}{\ding{55}}  & \textcolor{red}{\ding{55}} & \textcolor{red}{\ding{55}} & \textcolor{red}{\ding{55}} \\
  & JMLR & 17 & \textcolor{red}{\ding{55}} & 2 & 11 & \textcolor{red}{\ding{55}} & 2 & \textcolor{red}{\ding{55}} & \textcolor{red}{\ding{55}}  & \textcolor{red}{\ding{55}} & \textcolor{red}{\ding{55}} & \textcolor{red}{\ding{55}} \\
  & PMC\_LLaMA\_13B & 10 & \textcolor{red}{\ding{55}} & \textcolor{red}{\ding{55}} & 12 & 1 & 1 & \textcolor{red}{\ding{55}} & \textcolor{red}{\ding{55}} & \textcolor{red}{\ding{55}} & \textcolor{red}{\ding{55}} & \textcolor{red}{\ding{55}} \\
    \cmidrule(lr){2-13}
  & Med42-70B & 20 & \textcolor{red}{\ding{55}} & 11 & 25 & 2 & 2 & \textcolor{red}{\ding{55}} & \textcolor{red}{\ding{55}}  & \textcolor{red}{\ding{55}} & \textcolor{red}{\ding{55}} & \textcolor{red}{\ding{55}} \\
  & Apollo & 21 & \textcolor{red}{\ding{55}} & 5 & 24 & 1 & 2 & \textcolor{red}{\ding{55}} & \textcolor{red}{\ding{55}}  & \textcolor{red}{\ding{55}} & \textcolor{red}{\ding{55}} & \textcolor{red}{\ding{55}} \\
  & CancerLLM  & 16 & \textcolor{red}{\ding{55}} & 10 & 27 & 2 & 3 & \textcolor{red}{\ding{55}} & \textcolor{red}{\ding{55}}  & \textcolor{red}{\ding{55}} & \textcolor{red}{\ding{55}} & \textcolor{red}{\ding{55}} \\
  \bottomrule
\end{tabular}

\caption{
Performance of LLMs on Data-Driven and Knowledge-Driven Tasks under the zero-shot setting (eICU).
\textcolor{red}{\ding{55}} indicates no valid output.
\textcolor[HTML]{3166FF}{\ding{51}} indicates a perfect score of 100.
\textbf{\textcolor[HTML]{3166FF}{1\textsuperscript{st}}}, \textbf{\textcolor[HTML]{F56B00}{2\textsuperscript{nd}}}, and \textbf{\textcolor[HTML]{32CB00}{3\textsuperscript{rd}}} denote the best, second‑best, and third‑best results, respectively.
}
\label{tab:eicu_benchmark}
\end{table*}

\end{document}